\documentclass[conference, table]{IEEEtran}
\usepackage{graphicx}
\usepackage{mathtools}
\usepackage{amsfonts}
\usepackage{amssymb}
\usepackage{wrapfig}
\usepackage{times}
\usepackage{url}
\usepackage[margin=17pt]{subcaption}
\usepackage[final]{changes}
\usepackage[hidelinks]{hyperref}
\usepackage{xcolor}
\usepackage{colortbl}

\definecolor{LightCyan}{rgb}{0.9,1,1}
\definecolor{LightGray}{rgb}{0.95,0.95,0.95}

\newcommand{\ME}{\mathcal{E}}
\newcommand{\MV}{\mathcal{A}}
\newcommand{\MW}{\mathcal{W}}
\newcommand{\MT}{\mathcal{T}}
\newcommand{\ML}{L}
\newcommand{\BB}{\mathbb{B}}
\newcommand{\R}{\mathbb{R}}
\newcommand{\NN}{\mathbb{N}}

\newtheorem{definition}{Definition}[section]

\ifCLASSINFOpdf
  
\else
  
\fi

\begin{document}
\title{Explainable Predictive Process Monitoring}

\author{
    \IEEEauthorblockN{Riccardo Galanti\IEEEauthorrefmark{1}\IEEEauthorrefmark{3}, 
    Bernat Coma-Puig\IEEEauthorrefmark{2},
    Massimiliano de Leoni\IEEEauthorrefmark{3},
    Josep Carmona\IEEEauthorrefmark{2}, and
    Nicol\'o Navarin\IEEEauthorrefmark{3}}
    \IEEEauthorblockA{\IEEEauthorrefmark{1}myInvenio, Reggio Emilia, Italy,
    \IEEEauthorrefmark{3}University of Padua, Padua, Italy,
    \IEEEauthorrefmark{2}Universitat Politècnica de Catalunya, Barcelona, Spain
    \\
    Email: riccardo.galanti@my-invenio.com, 
    \{deleoni,nnavarin\}@math.unipd.it} 
    \{bcoma,jcarmona\}@cs.upc.edu}

\maketitle
\begin{abstract}
Predictive Business Process Monitoring is becoming an essential  aid for organizations, providing online operational support of their processes. 
This paper tackles the fundamental problem of equipping 
predictive business process monitoring with explanation capabilities, so that not only the {\em what} but also the {\em why} is reported when predicting generic KPIs like remaining time, or activity execution. We use the game theory of Shapley Values to obtain robust explanations of the predictions. The approach has been implemented and tested 
on real-life benchmarks,
showing for the first time how explanations can be given in the field of predictive business process monitoring.    
\end{abstract}

\IEEEpeerreviewmaketitle

\section{Introduction}
\label{sec:intro}

Within the field of Process Mining, predictive monitoring aims to forecast the running process instances with the purpose of timely signalling those that require special attention (those that may take too long, cost too 
much,
not be satisfactory, etc.). 
Several approaches have been proposed in literature
to deal with predictive monitoring (cf.\ Section~\ref{sec:related-work-prediction} and the survey by M{\'{a}}rquez et al.~\cite{Marquez-Chamorro18}), which has received significant attention in the last years. 
However, the majority of these approaches rely on black-box models (e.g.\ based on LSTM, i.e.\ Long Short-Term Memory neural models), which are proven to be more accurate, at the cost of being unable to provide a feedback to the user. On the other hand, approaches based on explicit rules (e.g.\ based on classification/regression trees) tend to be significantly less accurate. 
While the priority remains on giving accurate predictions, users need to be provided with an explanation of the reason why a given process execution is predicted to behave in a certain way.
Otherwise, users would not trust the model, and hence they would not adopt the predictive-monitoring technology~\cite{10.1007/s11257-017-9195-0,doshivelez2017rigorous}. 

 This paper tackles the problem of equipping process monitoring with explanations of the predictions. 
It leverages on current state of the art of Explainable AI (cf.\ Section~\ref{sec:related-work-explanation}), defining a framework for explainable process monitoring of generic KPIs. 
The proposed framework is independent of the machine- or deep-learning technique that is employed to make the predictions. However, we aim to instantiate the framework to prove its effectiveness. With this aim in mind, we built a process-monitoring framework, based on LSTM models, that is also able to explain any generic KPI, numerical or nominal. In a nutshell, given a running case, our framework estimates the future KPI value and returns the set of attributes that influence its prediction the most.

Experiments were conducted on different benchmarks, including the real-life process of an Italian financial institute, with the aim of predicting different KPIs, namely remaining time, costs, and the eventual occurrence of certain undesired activities. 
Explanations can be generated at LSTM-model level, to be provided to process stakeholders to understand the general trend of the model, but also at run-time, to explain the predictions of each single running case.
The explanations obtained for the aforementioned financial institute are in line with those of the analysts of the process. 
The remarkable difference is that our results were obtained within a few days of automatic computations, instead of long analyses. 

The rest of the paper is organized as follows.
Section~\ref{sec:probStatement} 
states the problem addressed in this paper.
Section~\ref{sec:related-work} summarizes the most relevant work related to process predictive monitoring and Explainable AI. Section~\ref{sec:LSTM} sketches the state of the art on using LSTM models for predictive monitoring, on which we build to provide explanations.
Section~\ref{sec:Explanation_KPI_Predictions} reports on our framework for explainable predictive process monitoring.
Section~\ref{sec:impl_&_experiments} reports on our framework's operationalization, and on the case studies conducted with an Italian financial institute, whereas Section~\ref{sec:conclusion} concludes the paper.

\section{Problem Statement}
\label{sec:probStatement}

The starting point for a prediction system is an \textit{event log}. 
An event log is a multiset of \emph{traces}. Each trace describes the life-cycle of a particular \emph{process instance} (i.e., a \emph{case}) in terms of the \emph{activities} executed and the process \emph{attributes} that are manipulated.
\begin{definition}[Events]
Let $\MV$ be the set of process attributes. Let $\MW_\MV$ be a function that assigns a domain $\MW_\MV(a)$ to each process attribute $a\in \MV$.
Let be $\overline{\MW} = \cup_{a \in \MV} \MW_\MV(a)$.
An \emph{event} $e \in \ME$ is a partial function $\MV \not\rightarrow \overline{\MW}$ assigning values to process attributes, with $e(a) \in \MW_\MV(a)$.
\end{definition}
Note that the same event can potentially occur in different traces, namely attributes are given the same assignment in different traces. 
This means that potentially the entire same trace can appear multiple times. This motivates why an event log is to be defined as a multiset of traces.\footnote{Given a set $X$, $\BB(X)$ indicates the set of all multisets with the elements in $X$.} 
\begin{definition}[Traces \& Event Logs]
Let $\ME$ be the universe of events.
A trace $\sigma$ is a sequence of events, i.e.\ $\sigma \in \ME^*$.
An event-log $\ML$ is a multiset of traces, i.e.\ $\ML \subset \BB(\ME^*)$.
\end{definition}
Predictive monitoring aims to estimate the future KPI values of the running cases. Here, we aim to be generic, meaning that KPIs can be of any nature:
\begin{definition}[KPI]
\label{def:KPI}
Let $\ME$ be the universe of events
defined over a set $\MV$ of attributes. Let $\MW_K$ be the domain of the KPI values.
A KPI is a function $\MT: \ME^* \times \NN \not\rightarrow \MW_K$ such that, given a trace $\sigma \in \ME^*$ and an integer index $i \leq |\sigma|$, $\MT(\sigma,i)$ returns the KPI value of $\sigma$ after the occurrence of the first $i$ events.\footnote{Given a sequence $X$, $|X|$ indicates the length of $X$. Notation $\not\rightarrow$ indicates that the function is partial.} 
\end{definition}
Note that our KPI definition assumes it to be computed a posteriori, when the execution is completed and leaves a complete trail as a certain trace $\sigma$. In many cases, the KPI value is updated after each activity execution, which is recorded as next event in trace; however, other times, this is only known after the completion. We aim to be generic and account for all relevant cases. Given a trace  $\sigma = \langle e_1,\ldots,e_n \rangle$ that records a complete process execution, the following are three potential KPI definitions:
\begin{LaTeXdescription}
\item [Remaining Time.]  $\MT_{remaining}(\sigma,i)$ is equal to the difference between the timestamp of $e_n$ and that of $e_i$. 
\item [Activity Occurrence.] It measures whether a certain activity is going to eventually occur in the future, such as an activity \emph{Open Loan} in a loan-application process. The corresponding KPI definition for the occurrence of an activity $A$ is $\MT_{occur\_A}(\sigma,i)$, which is equal to true if activity $A$ occurs in $\langle e_{i+1},\ldots,e_n \rangle$ and $i<n$; otherwise false.
\item [Customer Satisfaction.] This is a typical KPI for several service providers. Let us assume, without losing generality, to have a trace $\sigma = \langle e_1,\ldots,e_n \rangle$ where the satisfaction is known at the end, e.g.\ through a questionnaire. Assuming the satisfaction level is recorded with the last event - say $e_n(sat)$ . Then, $\MT_{cust\_satisf}(\sigma,i)=e_n(sat)$. 
\end{LaTeXdescription}
The following definition states the prediction problem:
\begin{definition}[The Prediction Problem]
Let $\ML$ be an event log that records the execution of a given process, for which a KPI $\MT$ is defined. Let $\sigma = \langle e_1,\ldots,e_k \rangle$ be the  trace of a running case, which eventually will complete as \linebreak \mbox{$\sigma_T=\langle e_1,\ldots,e_k, e_{k+1}\ldots,e_n \rangle$}.
The prediction problem can be formulated as forecasting the value of $\MT(\sigma_T,i)$ for all $k < i \leq n$. 
\end{definition}
As indicated in Section~\ref{sec:intro}, we aim to provide an explanation for the predictions. 
In particular, for each running case, we aim to return the set of attributes influencing its prediction the most, with the corresponding magnitude and the indication whether the attributes increase or decrease the predicted KPI's value.

In the light of the above, for each trace $\sigma = \langle e_1,\ldots,e_n \rangle$, the problem can be stated as finding a 
function $\mathcal{K}_{(\sigma,\MT)}$ such that, for all $a \in \MV$, $v \in \MW_\MV(a)$, and for all $i$ \mbox{s.t.\ $-n < i \leq 0$},  $\mathcal{K}_{(\sigma,\MT)}(a,v,i)$ is different from zero if and only if the assignment of value $v$ to attribute $a$ by $e_{(n-i)}$ has influenced the prediction of KPI $\MT$. The absolute value of $\mathcal{K}_{(\sigma,\MT)}(a,v,i)$ indicates how much this influence is, where a zero value indicates no influence. If $\mathcal{K}_{(\sigma,\MT)}(a,v,i) \neq 0$, its positive or negative sign indicates whether the influence is towards increasing or decreasing the KPI value:
\begin{definition}[The Prediction-Explanation Problem]
\label{def:pred-expl}
Let $\ML$ be an event log over a set $\MV$ of attributes, with domains $\MW_\MV$.
Let $\sigma = \langle e_1,\ldots,e_k \rangle$ be a running case with a KPI definition $\MT$. Let be $\overline{\MW} = \cup_{a \in \MV}~\MW_\MV(a)$.
Explaining the prediction is the problem of computing a 
function  \mbox{$\mathcal{K}_{(\sigma,\MT)}: \MV \times \overline{\MW} \times [-k+1,0] \rightarrow \R$}, where
$v \not\in \MW_\MV(a) \Rightarrow 
\mathcal{K}_{(\sigma,\MT)}(a,v,i)=0$.
\end{definition}

\section{Related Works}
\label{sec:related-work}

\subsection{Prediction of Process-Related KPIs}
\label{sec:related-work-prediction}

The predictive-monitoring survey of M{\'{a}}rquez et al.~\cite{Marquez-Chamorro18} reports on the large repertoire of techniques and tools that were developed to address this problem.
However, the authors claim that \textit{``little attention has been given to [$\ldots$] explaining the prediction values to the users so that they can determine the best way to act upon''}, and that \textit{``it is necessary to develop tools that help users to query these models in order to get information that is relevant for them''}. 
These are in fact the problems tackled in this paper, so as to ensure that the predictive-monitoring system is trusted, and thus used. 

Predictive monitoring has been built on different machine and deep-learning techniques, and also on their ensemble~\cite{Marquez-Chamorro18}. Different research works have recently illustrated that the so-called Long Short-Term Memory networks (LSTMs) generally outperform other methods (see, e.g., ~\cite{Park19,TaxVRD17,LSTM_time}).  
Therefore, while our explanation framework is independent of the machine- or deep-learning technique that is employed, we operationalize it with LSTMs. 
Section~\ref{sec:LSTM} provides further details on LSTMs, and details how they are employed for business-process predictive monitoring. 

It was explained above that little research work has been conducted on explaining the outcome of process predictive monitoring. The most relevant work is by Rehse et al.~\cite{Rehse2019}, which also aims at providing a dashboard to process participants with predictions and their explanation. However, the paper does not provide sufficient details on the actual usage of the explainable-AI literature, and the very preliminary evaluation is based on one single artificial process that consists of a sequence of five activities. Breuker et al. also try to tackle the problem~\cite{10.1007/978-3-319-15895-2_46}, but their attempt is not independent of the actual technique employed for predictions. Furthermore, their explanations are only based on the activity names, while the explanations can generally involve resources, time, and more (cf.\ the case studies reported in Section~\ref{sec:impl_&_experiments}).

\subsection{Explanation of Machine-Learning Models}
\label{sec:related-work-explanation}
Few approaches exist in the literature to explain machine learning models, arisen from the need to understand complex black-box algorithms like ensembles of Decision Trees and Deep Learning~\cite{lime,shrikumar2017learning,gradients,shap}.

The adoption of explanatory methods in industry is at an early stage;
in \cite{Shu:2019:DEF:3292500.3330935} an approach of fake news detection grounded in explainability is introduced.
A significant amount of work in literature is focused on healthcare applications.
We highlight  \cite{lundberg2018explainable}, an implementation of the Shapley Values in healthcare, where the explanatory method is used to prevent hypoxaemia during surgery, and
\cite{inbook}, where explainability is used for analysis of patience re-admittance.

The SHAP implementation of the Shapley values for Deep Learning has the strong theoretical foundation of the original game theory approach, with the advantage of providing offline explanations that are consistent with the online explanations. Moreover, SHAP avoids the problems in consistency seen in other explanatory approaches (e.g. the lack of robustness seen in the online surrogate models, as analysed in \cite{alvarez2018robustness}). The framework proposed in this paper specializes the use of Shapley values to the problem of providing explanations for predictive analytics. 

We also considered attention mechanisms~\cite{DBLP:conf/iclr/2015} as an alternative. However, two limitations made us opt for Shapley values. First, attention mechanisms necessarily have to be integrated in a Neural Network architecture,
while Shapley values can be applied to any Machine or Deep Learning algorithm. The second limitation is linked to the lack of consensus that attention weights are always correlated to feature importance. Jain et al.~\cite{DBLP:conf/naacl/JainW19} find it \textit{``at best, questionable – especially when a complex
encoder is used, which may entangle inputs in the hidden space''},
Serrano et al.~\cite{serrano-smith-2019-attention} state that
\textit{``attention  weights  often  fail  to  identify the sets of representations most important to the  model’s  final  decision''}.

\section{The Use of LSTMs for Predictive Monitoring}
\label{sec:LSTM}
As indicated in Section~\ref{sec:intro}, we implemented our framework by leveraging on LSTM models~\cite{Hochreiter1997}, a special type of Recurrent Neural Networks. LSTM models natively support the predictions where the independent variables are sequences of elements, and the literature has shown that they are among the most suitable methods for predictive business monitoring (cf.\ Section~\ref{sec:related-work-prediction}). 

The construction of LSTM models fall into the problem of supervised learning, which aims to learn the model from a training set, for which the value of the dependent variable is known. This set is composed by pairs $(x,y) \in \mathcal{X} \times \mathcal{Y}$ where $\mathcal{X}$ represents the independent variables with their values (also known as \textbf{features}), and $\mathcal{Y}$ is the value observed for the dependent variable (i.e.\ the value we aim to predict).

In the domain of LSTM learning, $\mathcal{X}$ consists of sequences of vectors with a certain number $n$ of dimensions, i.e.\ $\mathcal{X}={(\R^n)}^*$.\footnote{In literature, LSTMs are often trained on the basis of matrices. However, a sequence of $m$ vectors in $\R^n$ can be seen, in fact, as a matrix in $\R^{n,m}$. We use here the dataset representation as vectors to simplify the formalization.}
When LSTM is used for predictive business monitoring using KPI values in a domain $\MW_K$ (cf.\ Definition~\ref{def:KPI}), $\mathcal{Y}$ is $\MW_K$. 

With these preliminaries at hand, we built a process monitoring framework composed by two phases: off-line and on-line.

The off-line phase requires an event log $\ML$ and a KPI definition $\overline{\MT}$ as input. This enables creating the dataset for training and testing the LSTM model, which consists of pairs $(x,y) \in {(\R^n)}^* \times \MW_K$. The input is, hence, a sequence of vectors; conversely, a trace is a sequence of events. Therefore, each event needs to be encoded as a vector, 
which is a problem largely studied: we use the same encoding as in \cite{LSTM_time}; this can be abstracted as an \textbf{event-to-vector encoding function} $\rho: \ME \rightarrow \R^n$. In a nutshell, each  numeric attribute $a$ of event $e$ becomes a different dimension of $\rho(e)$, which takes on value $e(a)$. Each boolean attribute $a$ is also a different dimension, with either $0$ or $1$ depending whether $e(a)$ is false or true. Each literal attributes $a$ is represented through the so-called \textit{one-hot encoding}: one different dimension exists for each value $v \in \MW_\MV(a)$, and the dimension referring to value $e(a)$ takes on value $1$, with the other dimensions be assigned value $0$.
Function $\rho$ can also be overloaded to traces: $\rho(\langle e_1, \ldots, e_m \rangle)=\langle \rho(e_1), \ldots, \rho(e_m) \rangle$.

The dataset is created starting from each prefix $\sigma'$ of each trace $\sigma \in \ML$: $\sigma'$ will generate one item in the data set consisting of a pair $(x,y) \in {(\R^n)}^* \times \MW_K$
 where $x=\rho(\sigma')$ and $y=\overline{\MT}(\sigma,|\sigma'|)$.
The dataset is later divided in one larger part for training the LSTM model, and a smaller part for testing. The test part is used to evaluate the quality of the LSTM model, in terms of different metrics. Details of the proportions and the quality metrics employed are discussed in Section~\ref{sec:impl_&_experiments}.
The \textbf{LSTM-based process predictor} trained from a dataset $\mathcal{D} \subset {(\R^n)}^* \times \MW_K$ can be abstracted as a function $\Phi_\mathcal{D}: \mathbb{R}^{n^*}\rightarrow \MW_K$.

The on-line phase aim is to predict the KPI of interest for a set of running cases of the process, identified by a set $\ML'$ of partial traces (i.e.,\ a log). It relies on the LSTM-based process predictor $\Phi_\mathcal{D}$: for each $\sigma'  \in \ML'$, the predicted KPI value is $\Phi_\mathcal{D}(\rho(\sigma'))$.

\section{Explanation of Generic KPI Predictions}
\label{sec:Explanation_KPI_Predictions}

This section reports on the main contribution of this paper, namely using Shapley Values to explain the predictions of any predictive model. 

Section~\ref{sec:shapTheory} introduces the theory behind Shapley values, while Section~\ref{sec:sv} illustrates its application and adaptation for predictive process monitoring. Then, in Section~\ref{sec:expl_approach} we provide the general picture and the two main types of explanations reported.

\subsection{The Theory of Shapley Values}
\label{sec:shapTheory}

The Shapley Values \cite{shapley1953value} is a game theory approach to fairly distribute the payout among the players that have collaborated in a cooperative game. This theory can be adapted as an approach to explain a predictive model. 
The assumption is that the features from an instance correspond to the players, and the payout is the difference between the prediction made by the predictive model and the average prediction (later referred to as the {\em base value}). 
Intuitively, given a predicted instance, the Shapley Value  of a feature expresses how much the feature value contributes to the model prediction~\cite{molnar2019}: 
\begin{definition}[Shapley Value]
Let $X=\{x_1,\ldots,x_n\}$ be a set of features. The Shapley value for feature $x_i$ is defined as:
\begin{equation*}
\resizebox{\columnwidth}{!}{$
   \psi_i=\sum_{S\subseteq\{ x_1, \ldots, x_m \}\setminus\{x_i\}}\frac{|S|!\left(p-|S|-1\right)!}{p!}\left(val\left(S\cup\{x_i\}\right)-val(S)\right)
   $}
\end{equation*}
where $val(T)$ is the so-called payout for only using the set of feature values in $T \subset X$ in making the prediction. 
 \label{def:SV}
\end{definition}

Intuitively, the formula in Definition~\ref{def:SV} evaluates the effect of incorporating the feature value $x_i$ into any possible subset of the feature values considered for prediction. In the equation, variable $S$ runs over all possible subsets of feature values, the term $val\left(S\cup\{x_i\}\right)-val(S)$ 
corresponds to the marginal value of adding $x_i$ in the prediction using only the set of feature values in $S$,
and the term
$\frac{|S|!\left(p-|S|-1\right)!}{p!}$ corresponds to all the possible permutations with subset size $|S|$, to weight different sets differently in the formula.
This way, all possible subsets of attributes are considered, and the corresponding effect is used to compute the Shapley Value of $x_i$. 

\subsection{Explainable Predictions through Shapley Values}
\label{sec:sv}
The starting point is a event-to-vector encoding function $\rho: \ME \rightarrow \R^n$ that maps each event to a feature vector (cf.\ Section~\ref{sec:LSTM}). 
Given an event $e_i$, $\rho(e_i)=[x_i^1, \ldots,x_i^n]$ where each feature $x_i^j$ is associated with an event attribute $a_i^j$ and, possibly, with a value $v_i^j$. We mentioned that, if an attribute $a_i^j$ is categorical, we need to introduce as many features as its possible values (one-hot encoding).
Namely, $x_i^j$ is both associated with an attribute $a_i^j$, and with a value $v_i^j$. If the feature associated with attribute $a_i^j$ and value $v_i^j$ takes on value $1$, then $e(a_i^j)=v_i^j$; otherwise, the value is $0$. If an attribute $a_i^j$ is conversely numerical, only one feature $x_i^j$ exists with value $e(a_i^j)$.
When applied for explainable predictive monitoring, the Shapley values of a trace $\sigma=\langle e_1, \ldots, e_m \rangle$ are computed over the features of the vector $\chi$ =
$[x_1^1, \ldots, x_1^n,\ldots,x_m^1, \ldots, x_m^n]$ where
$\rho(e_i)=[x_i^1, \ldots, x_i^n]$ for $1 \leq i \leq m$.

When applying Definition~\ref{def:SV} to all features of $\chi$, the result is a vector of Shapley values $\Psi=[\psi_1^1, \ldots, \psi_1^n, \ldots, \psi_m^1, \ldots ,\psi_m^n]$ associated to feature vector $\chi$, and attributes $[a_1^1, \ldots, a_1^n, \ldots, a_m^1, \ldots, a_m^n]$. 
Any Shapley value $\psi_i^j$ can be either positive or negative. A positive or negative value indicates that the feature contributes to increasing or decreasing the value, respectively.

This allows us to construct the explanations. The first step is to determine which features are relevant and at which timestep.\footnote{In this context each timestep refers to a different event of the trace along with its attributes (features). For instance, timestep zero refers to the first event of the trace, timestep one to the second, etc.} For this, we consider the average $\mu$ of the values in $\Psi$ along with their standard deviation $\xi$.
This allows to define an interval $I=[\mu-\delta \xi, \mu+ \delta \xi]$ of Shapley values that are not considered to contribute significantly, where $\delta$ is a parameter set by the user. This reduces the number of features that are considered in the explanation, limiting its verbosity.

Let us consider each Shapley value $\psi_i^j \not\in I$, associated with feature $x_i^j$ and an event's attribute $a_i^j$. 

\textbf{If $a_i^j$ is a numerical attribute}, attribute $a_i^j$ is the explanation itself, i.e.\ $\forall \overline{v} \in \MW_\MV(a).$ $\mathcal{K}_{(\sigma,\MT)}(a_i^j,\overline{v},i-m)=\psi_i^j$. 

\textbf{If $a_i^j$ is a categorical attribute}, $x_i^j$ is a one-hot encoded feature, and it is also associated with a value $v_i^j$. If $x_i^j=1$, the explanation obtained is that $a_i^j = v_i^j$ contributes to the KPI value: 
$\mathcal{K}_{(\sigma,\MT)}(a_i^j,v_i^j,i-m)=\psi_i^j$. Otherwise, $x_i^j=0$, and the explanation is $a_i^j \neq v_i^j$, namely $\forall \bar{v} \in \MW_\MV(a) \setminus \{v_i^j\}.$
$\mathcal{K}_{(\sigma,\MT)}(a_i^j,\bar{v},i-m)=\psi_i^j$.

Any other combination $(a,v,i)$ that does not fall into the situations above is such that $\mathcal{K}_{(\sigma,\MT)}(a,v,i)=0$.

\begin{figure}[t]
    \centering
    \includegraphics[width=1 \columnwidth]{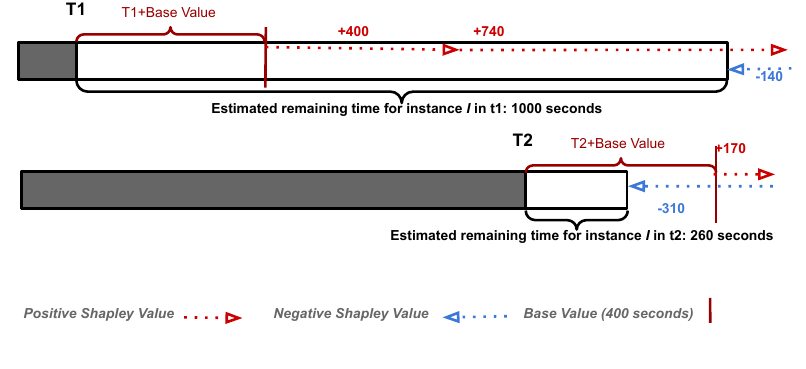}
    \caption{Two explanations examples using Shapley Values. When the Remaining Time predicted is high (i.e. higher than Base Value), the Shapley Values indicate which features increase the prediction. Similarly, when the prediction is smaller than the Base Value, most of the Shapley Values are negative. }
    \label{fig:TimeConsiderations}
\end{figure}

While an exact computation of the Shapley values requires to consider all combinations of features (hence, the algorithm is exponential on the number of features), efficient estimations can be obtained through polynomial algorithms that use greedy approaches~\cite{molnar2019}.

To conclude, let us illustrate how Shapley values help explain a typical KPI in predictive process monitoring: estimated remaining time.
Figure \ref{fig:TimeConsiderations} shows the estimated remaining time of the same case in two different moments: T1, when the case started (upper figure, with an estimated remaining time of 1000 seconds), and T2, when it is close to its end (lower figure, with an estimated remaining time of 260 seconds). Considering that 
the Base Value is 400 seconds, the explanatory method would indicate, at T1, which features have been useful for the predictive model to predict a high value, i.e. the features with a positive Shapley Value. On the other hand, for T2, most of the Shapley Values would be negative, since the model has predicted a value smaller than the base value.

\subsection{Overall Approach for Explaining Generic KPI Predictions}
\label{sec:expl_approach}

\begin{table*}[t!]
\caption{Online explanations for \emph{Remaining Time} for three running cases. When the explanation is followed by $(-1)$, it means that it refers to the value assigned to the attribute by the event that precedes the last of respective case.}
\footnotesize
\centering
\resizebox{\textwidth}{!}
{\begin{tabular}{|l|l|p{6cm}|l|}
\hline
\cellcolor{LightCyan}\textbf{CASE ID} & \scriptsize \cellcolor{LightCyan}\textbf{REMAINING TIME} & \scriptsize \scriptsize \cellcolor{LightCyan}\textbf{Explanations for increasing remaining\_time}  &\cellcolor{LightCyan}\textbf{Explanations for decreasing remaining\_time} \\
\hline
\scriptsize201810011258   & \scriptsize5d 6h 7m  & 
\scriptsize ACTIVITY=Evaluating Request (NO registered letter)   &\scriptsize CLOSURE\_TYPE!=Inheritance    \\

\scriptsize\cellcolor{LightGray}201810000206  & \scriptsize \cellcolor{LightGray}5d 2h 12m  & 
\scriptsize\cellcolor{LightGray} ROLE=DIRECTOR   &\scriptsize\cellcolor{LightGray}\scriptsize CLOSURE\_TYPE=Bank Recess    \\
\scriptsize201811010829  & \scriptsize2d 2h 31m  &\scriptsize ROLE!=BACK-OFFICE (-1) AND ACTIVITY!=Service closure Request with BO responsibility (-1) & \scriptsize- \\ 
\scriptsize\cellcolor{LightGray}...     & 
\scriptsize\cellcolor{LightGray}...    & 
\scriptsize\cellcolor{LightGray}...  & 
\scriptsize\cellcolor{LightGray}...    \\                \hline     
\end{tabular}}
\label{tab:rem_time_online_explanations}
\end{table*}

Explanations can be used offline to explain the features/factors that the trained model uses to make predictions, moreover they can be employed online on each running case to put forward the factors that affected the predictions. In particular, offline explanations are calculated on the test dataset, which is a part of the dataset not used for training the model (information about the division between train and test sets will be provided in Section~\ref{sec:impl_&_experiments}).

\subsubsection{\textbf{Offline Explanations}}

Our offline explanation strategy is to provide an heatmap that overviews the importance of each factor in explaining the instances of the test dataset.

\begin{figure}[t!]
    \includegraphics[width=1 \columnwidth]{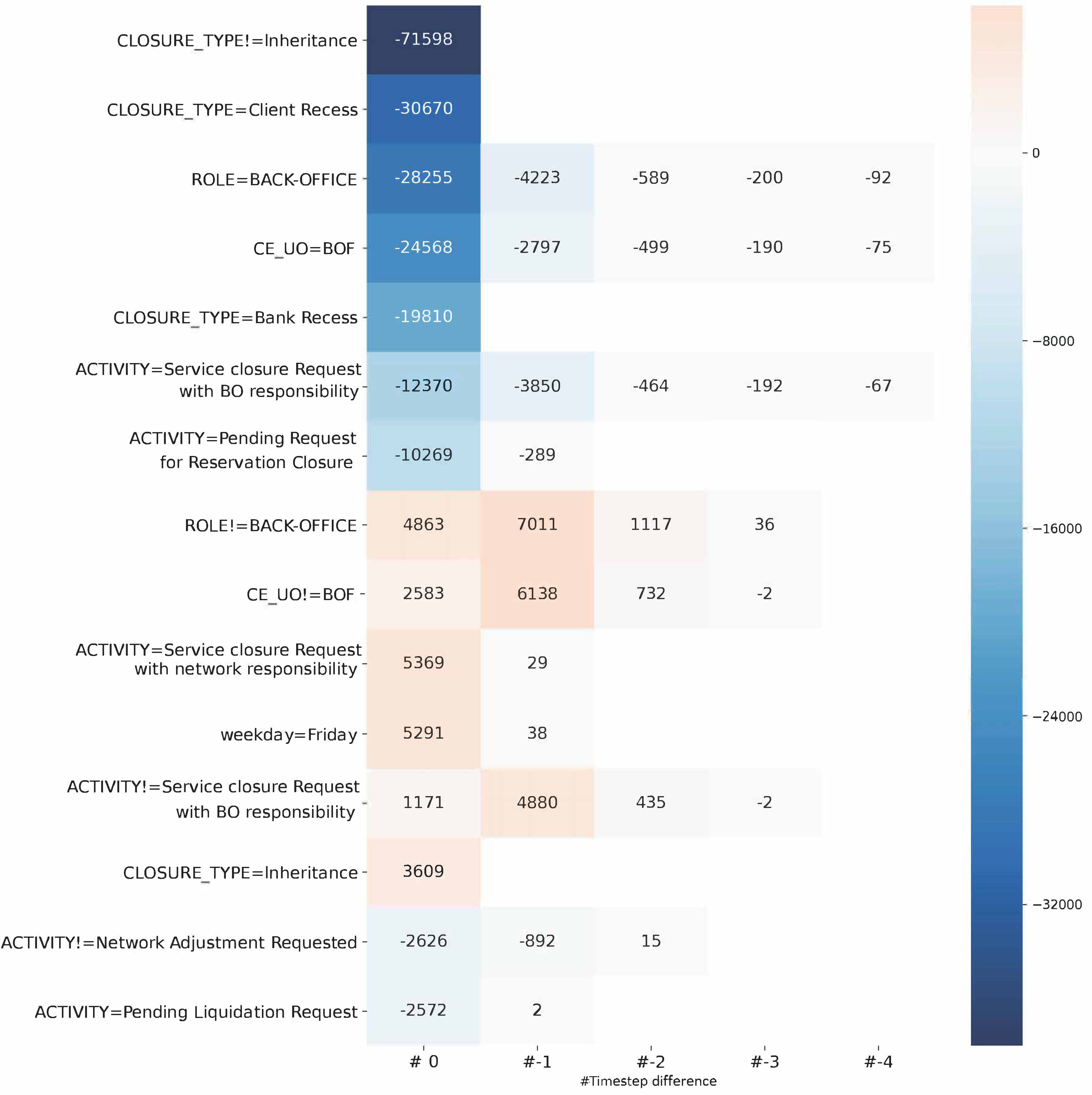}
    \caption{The offline explanation of the remaining time}
    \label{fig:explanation_rem_time}
\end{figure}

In particular, given an event log $L$, we consider each prefix $\sigma'$ of each trace in $L$.
Then, we compute the explanations as defined in Section~\ref{sec:sv}. Figure~\ref{fig:explanation_rem_time} shows an example of a heatmap reporting the frequency in which an explanation is relevant after each event in every trace.
The $y$ axis lists different explanations of types \mbox{$attr=value$} or $attr \neq value$ while the $x$ axis 
lists the timestep difference between the event in question and
the last event of the considered prefix, namely 0 indicates the last event, -1 indicates the second last, etc.
A cell with explanation $a = v$  (y axis) and timestep difference $t$ (x axis) takes on a value $(x-y)$ if there are $x$ prefixes $\sigma'$ of traces in $L$ s.t.\ $\mathcal{K}_{(\sigma',\MT)}(a,v,t)>0$ and $y$ prefixes $\sigma''$ of traces in $L$ s.t.\ $\mathcal{K}_{(\sigma'',\MT)}(a,v,t)<0$. For instance, let us consider the explanation \texttt{ROLE=BACK-OFFICE} with timestep difference 0, which is associated with value $-28255$. This means that $-28255$ is the difference between the number of prefixes $\sigma'$ in which \texttt{ROLE=BACK-OFFICE} in the last event of $\sigma'$ contributes to increasing the KPI value and the number of prefixes $\sigma'$ in which \texttt{ROLE=BACK-OFFICE} of last event contributes to decreasing. Similarly, $-4223$ is the difference when considering the second last event of the prefixes in place of the last.
A similar reasoning can be repeated for explanations of type $a \neq v$. 
The heatmap uses different shades of blue and red to highlight the magnitude of negative and positive values, respectively. 

\subsubsection{\textbf{Online Explanations}} 
When we focus on running cases, we generate a table with one row per running case (see, e.g., Table~\ref{tab:rem_time_online_explanations}. Each row shows the case id, unique for each running case, the prediction for the current KPI, and the explanations that influence the prediction.
Section~\ref{sec:impl_&_experiments} discusses the case study in detail, including the results in Table~\ref{tab:rem_time_online_explanations}.

\section{Implementation and Experiments}
\label{sec:impl_&_experiments}

The framework for explainable predictive monitoring has been implemented in Python, using Pandas to elaborate the data, and the shap library\footnote{\url{https://shap.readthedocs.io/en/latest}} to explain the prediction.\footnote{Code can be found at \url{https://github.com/PyRicky/LSTM_Generic_explainable}} We relied on Keras framework for the LSTM implementation. The architecture was composed by 8 layers with 100 neurons each.

Each LSTM model was trained in 12-24 hours, and the computation of the off-line explanations (i.e. the heatmaps) required a similar amount of time. For each running case, on-line predictions and explanations are given in ca.\ half second. Note that training models in less than one day does not pose significant limitations: this is just performed once before putting the system in production.

The remainder of the paper will report on the experiments with different KPIs for the process carried on in an Italian bank. 
However, we also conducted several additional experiments with publicly-available event logs, which confirm the findings reported here.
Space limitation prevent us from reporting on them, which are however discussed in the appendix that complements this paper.

\subsection{Domain description}
\label{sec:Dataset}
Our assessment is based on the so-called \textbf{\emph{Bank Account Closure}}, a process executed at an Italian Banking Institution. The process deals with the closure of customer's accounts, which may be requested either by the customer or by the bank, for several reasons.

From the bank's information system, we extracted an event log with 32.429 completed traces and 212.721 events. It contains 15 different activities, 654 possible resources (recorded in an attribute labeled \emph{Ce\_Uo}), divided in 3 roles (attribute \emph{role}). 
Each trace is associated with an attribute \emph{Closure\_Type}, which encodes the type of procedure that is carried out for the specific account holder, and the \emph{Closure\_Reason}, namely the reason triggering the closure's request. The latter is only known for 79.43\% of cases. 

For the bank, it is of interest to obtain an estimate of the remaining time until the end for running cases. This allows the bank to decide which cases require special attention, in order to not postpone them too much further. Also, the bank wants to be informed whether there are high chances that one or more of the following activities will occur: \emph{Authorization Requested}, \emph{Pending Request for Acquittance of heirs}, and \emph{Back-Office Adjustment Requested}. They are linked to contingency actions, which should be avoided because they would cause inefficiencies in terms of time, costs, and resource utilization. Finally, the bank is also interested in obtaining an estimate of the total cost of a running case, in order to detect in advance which cases require particular attention.

We used two/thirds of the traces as training, and one third as test set.
For improving the quality of the trained model, we used hyperparameter optimization, with 20\% of the training data employed for this (validation set).

Sections~\ref{sec:rem_time_prediction}, \ref{sec:activity_prediction} and \ref{sec:cost_prediction} report on the outcome for remaining-time prediction, for the prediction of the occurrence of one of those three contingency actions and for total cost prediction, respectively.

\subsection{Results on Remaining Time}
\label{sec:rem_time_prediction}

Section \ref{sec:Explanation_KPI_Predictions} showed that the explanation for a learnt prediction model is given as a heatmap during the offline phase.
Figure \ref{fig:explanation_rem_time} refers to the application for the remaining time prediction. The fact that the closure type is not Inheritance (\emph{Closure\_Type!=Inheritance}) is the largest value in the heatmap (as absolute value), so it is the largest factor that influences the prediction. The information that the value is negative (i.e. -71598) indicates that the influence is towards reducing the value, namely towards having lower remaining time.
From a domain viewpoint, when the type of procedure is Inheritance, the bank-account holder is passed away. A further analysis of the data confirms this finding: if the type is \emph{Inheritance}, the process duration is 29 days, versus 14 days when the type is different. The evidence in the explanation illustrates that LSTM allowed learning a prediction model that leverages on the closure type to estimate the remaining time.
Other important attributes are related to the role associated to the resource and the resource performing each activity. Let us consider attributes \emph{Role=Back-office} and \emph{CE\_UO=BOF} that are related to back-office activities, which are generally performed in the final part of cases; it can be seen in the heatmap that even in this case the model is able to predict that the process instance is about to complete (a negative value again indicates smaller remaining time).

The discussion was so far focused on the attribute of the last event. However, the values of attributes of previous events also influence the prediction of remaining time as shown in the heatmaps (see columns related to timestep differences -1, -2, -3 and -4). Consider, e.g., the row \emph{ROLE=Back-office} and column -1: the value -4223 indicates that if the previous event refers to an activity performed by a resource with role Back-office, this influences to lower the prediction: the case is getting even closer to the end.
When activities are performed by a resource director the behaviour is considered as exceptional, while activities performed by resources playing the role of applicant are in general performed in the initial part of cases; consequently, the cases usually take longer to complete. This is indicated by the positive value 4863 of the last event in the row \emph{ROLE!=Back-office}, which indicates that the influence is towards increasing the remaining time. Notice that the column related to timestep difference -1 has a bigger value (7011), indicating that if the previous event refers to an exceptional activity, the influence on the prediction will be even stronger.
Finally when the activity performed is other than Network Adjustment Requested then the predicted remaining time is smaller; this is in fact an exceptional activity, that only occurs when an error is made in the early stages of the process, and even in this case our framework was able to learn to predict a smaller remaining time when no adjustments need to be done.

Section V indicated that explanations are also given for running cases to explain predictions to process stakeholders. Our implementation returns a CSV file with the predictions for the running cases; a subset is provided in Table \ref{tab:rem_time_online_explanations}, which shows the factors that increase or decrease the prediction for the remaining time prediction. 
Let us consider as an example the last row: the remaining time is predicted as being ca. 2 days and 2 hours, with two explanations increasing the prediction, one related to the fact that the previous activity performed was not \textit{Service Closure Request with BO Responsibility}, and the other related to the resource performing the previous activity with a role not being  \textit{Back-Office}.

To conclude, since this KPI is numerical and the values are reasonably well balanced, we adopted the \emph{Mean Absolute Error} (MAE), which is the average difference between the actual and the predicted value, computed over all test-set samples. Here, we achieved a MAE of 4.37 days, which is around the 28\% of the average case duration (i.e.\  15.5 days).

\subsection{Results on Prediction of Activity Occurrence}
\label{sec:activity_prediction}

\begin{table*}[t!]
\caption{Online explanations for \emph{Back-Office Adjustment Requested}. Values 1 and 0 indicate if the activity is predicted to occur or not.
Explanation followed by $(-1)$: attribute value assigned by the event that precedes the last of respective case.}
\footnotesize
\centering
\resizebox{\textwidth}{!}
{\begin{tabular}{|l|p{3cm}|p{5cm}|p{5cm}|}
\hline
\cellcolor{LightCyan}\textbf{CASE ID} &
  \cellcolor{LightCyan}\textbf{Back-Office Adjustment Requested} &
  \cellcolor{LightCyan}\textbf{Explanations for Back-Office Adjustment  Requested happening} &
  \cellcolor{LightCyan}\textbf{Explanations for Back-Office Adjustment Requested not happening} \\
  \hline
\scriptsize201810000206 &
  \scriptsize0 &
  \scriptsize- &
  \scriptsize{ACTIVITY=Service closure Request with network responsibility (-2) AND CE\_UO=195 (-1)} \\
\scriptsize \cellcolor{LightGray}201811008237 &
  \scriptsize \cellcolor{LightGray}1 &
  \scriptsize\cellcolor{LightGray}{CLOSURE\_TYPE=Porting} &
  \scriptsize \cellcolor{LightGray}- \\
\scriptsize201812005701 &
  \scriptsize1 &
  \scriptsize{CLOSURE\_REASON!=1 - Client lost} &
  \scriptsize- \\
\scriptsize \cellcolor{LightGray}... &
  \scriptsize \cellcolor{LightGray}... &
  \scriptsize \cellcolor{LightGray}... &
  \scriptsize \cellcolor{LightGray}...\\
  \hline
\end{tabular}}
\label{tab:bo_adj_online_explanations}
\end{table*}

We mentioned that the financial institute aims to avoid activities related to inefficiencies (e.g. rework): \textit{Pending Request for Acquittance of Heirs}, \textit{Back-Office Adjust Requested} and \textit{Autorization Required}.
Space limitation prevents us from showing here all of three: here we focus on activity \emph{Back-Office Adjust Requested}, while the other two are in the appendix complementing this paper.
The learnt LSTM model was characterized by an F1 score of 0.65, an Area Under the Receiver Operating Charateristics (AUROC) of 0.86, and an Area under Precision/Recall curve (APR) of 0.69. We computed AUROC and APR, because these metrics are, in fact, more suitable when some classes are unbalanced. This is actually the case for our case study: the three activities are contingency actions, which occur infrequently.
\begin{figure}[ht]
    \includegraphics[width=1 \columnwidth]{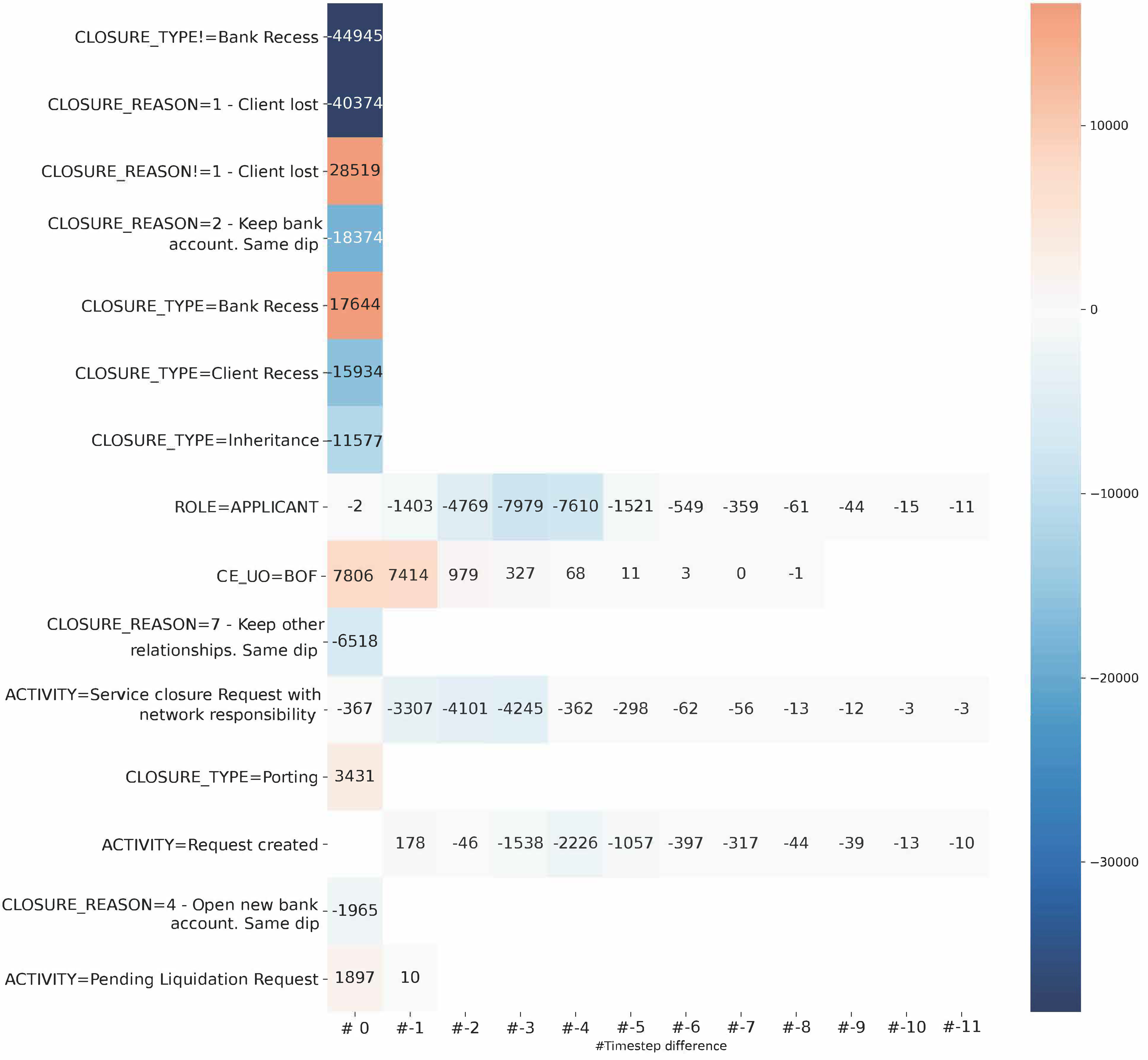}
    \caption{Offline explanations for \emph{Back-Office Adjustment Requested}}
    \label{fig:BO_adjustment}
\end{figure}

The heatmap related to \emph{Back-Office Adjustment Requested} prediction (Figure \ref{fig:BO_adjustment}) shows that the attributes related to the type and the reason of bank account closure are influencing the most. 
When all bank accounts of a customer are closed (labeled by \emph{Closure\_Reason=1 - Client lost}) or when the customer decides to close one of its bank accounts among different ones he owns (labeled as \emph{Closure\_Reason=2 - Keep bank account. Same dip}), then a \emph{Back-Office Adjustment Requested} is unlikely to happen. This is clearly shown in the heatmap, respectively represented by the values -40374 and -18374, which influence is towards not predicting the occurrence of this activity. 
Values -15934 when the Closure Type is Client Recess (it is the client that decides to close the bank account) and -11577 when it is Inheritance (the bank-account holder is passed away) indicate as well that a \emph{Back-Office Adjustment Requested} is unlikely to happen.
Conversely, when the Closure Type is Bank Recess (the bank account is closed by the bank) or it is Porting, then the rework activity \emph{Back-Office Adjustment Requested} is more likely to occur.

Explanations are also used on-line to explain the predictions of running cases. Table \ref{tab:bo_adj_online_explanations} shows the factors that make the model predict whether or not activity \emph{Back-Office Adjustment Requested} is expected to happen for three running cases. 
Values 1 and 0 indicate that the activity is expected or not to happen, respectively.
Let us consider for instance the first case in the table: the rework activity is not expected to happen because two events ago \textit{Service Closure Request with Network Responsibility} has been performed and because the previous event has been performed by the resource 195.
Conversely, it is predicted to eventually happen for the other two cases in the table, and the explanation is related to the closure type being Porting and the closure reason not being Client lost. 

\begin{figure}[t]
    \includegraphics[width=1 \columnwidth]{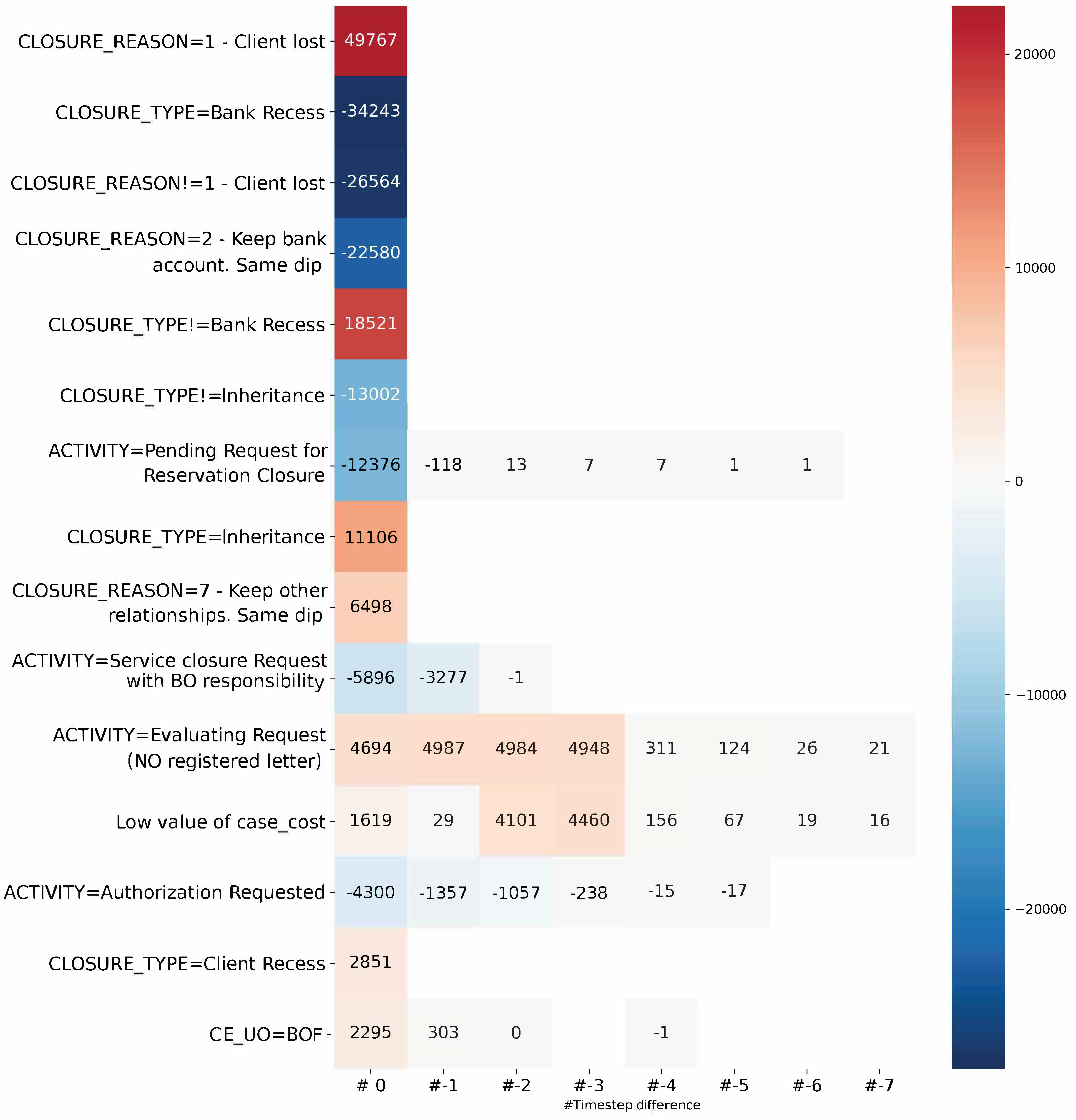}
    \caption{Offline explanations for \emph{Case cost}}
    \label{fig:Case_cost}
\end{figure}

\subsection{Results on Case Cost prediction}
\label{sec:cost_prediction}

Since this KPI is numerical, we adopted the \emph{Mean Absolute Error}, for which we achieved a value of 0.95 Euros. This is an excellent result, given that the average case cost is 12.86, with standard deviation of 6.41.
Figure \ref{fig:Case_cost} shows the application for the case cost prediction for the off-line phase.
The main factor that contributes to increase the cost of a case is represented by \emph{Closure\_Reason=1 - Client Lost}, which is indicated when all bank accounts are going to be closed. The information that the value is positive (i.e. 49767)  indicates  that  the  influence  is  towards  increasing  the cost. 
This is mainly caused by the fact that most of the times here the director needs to carefully evaluate the request before proceeding, and the hourly director's wage is certainly higher than that of other bank employees. 
Nevertheless, this evaluation is not needed when the closure of the bank account is requested by the bank (labeled as \emph{Bank Recess}), therefore the predicted case cost will be smaller (indicated in the heatmap by the negative value -34243). The director is similarly not involved when  customers only close one of their bank accounts (\emph{Closure\_Reason=2 - Keep bank account. Same dip}), which is a factor that yields lower costs.
Another reason is that when only one between different bank accounts is closed, then of course the process is simpler and less Back-office adjustment activities need to be performed compared to when all bank accounts need to be closed, leading to minor costs.
Another indirect evidence that the director's involvement is a factor that increases costs is evident when one looks at the explanations based on \emph{Activity=Evaluating Request (NO registered letter)}. This activity needs a lot of time and is performed by the director, leading to high costs (even higher compared to the case in which a request has only to be authorized). If this activity occurs, the cost will remain permanently high. This is evident in the heatmaps: the fact that this activity has been previously performed 
is still influencing towards increasing the costs (see columns related to timestep difference -1, -2 and -3, which values are respectively 4987, 4984 and 4948).

\section{Conclusion}
\label{sec:conclusion}
A lot of research has been devoted towards increasingly accurate frameworks for predictive process monitoring. Nonetheless, little attention has been paid to ensure that that the resulting predictive-monitoring system is workable in practice. With practical workability, here we intend that the process analysts and stakeholders need to trust the system and its predictions. Previous studies have shown that a necessary condition to build trust is to explain the reason of the provided predictions~\cite{10.1007/s11257-017-9195-0,doshivelez2017rigorous}. Proposals that do not put explanation as a core feature are not going to be adopted in practice.  

This paper has put forward a framework to equip predictive-process-monitoring systems with explanations that are intelligible by actors of the process. The framework builds on the most recent state of the art on Explainable AI, and is independent of the actual AI 
predictive-analytics technique.

However, the operationalization of the framework requires one to select an actual AI technique, and here we opted for predictive models based on LSTM, which the present literature has shown to be the most suitable for the problem in question. The implementation is based on Python, and it has been used for several case studies. Here we reported different KPI predictions for a process run in a financial institute in Italy. The case studies shows that our framework is able to, on the one hand, provide explanations of the most salient features that influence the prediction models and, on the other hand, to provide online explanations on the running cases. 

Future work accounts different directions. First, we aim to verify through interviews whether process stakeholders  would fully comprehend the heatmaps and the form given to explanations. Second, we aim to explore the possibilities
of {\em Natural Language Generation} techniques to report more user-friendly explanations, instead of the output shown in Tables~\ref{tab:rem_time_online_explanations} and \ref{tab:bo_adj_online_explanations}.

\appendix
\label{appendix}

This section is intended to briefly show and explain additional experiments that have been made on several public datasets, in order to show the general validity of our framework. The characteristics of the events logs are presented in Table~\ref{tab:dataset_statistics}, while the results obtained on those datasets are shown in  Table~\ref{tab:results}.

\subsection{Bank Account Closure}

\begin{figure*}[h!]
    \centering
    \includegraphics[width=1 \textwidth]{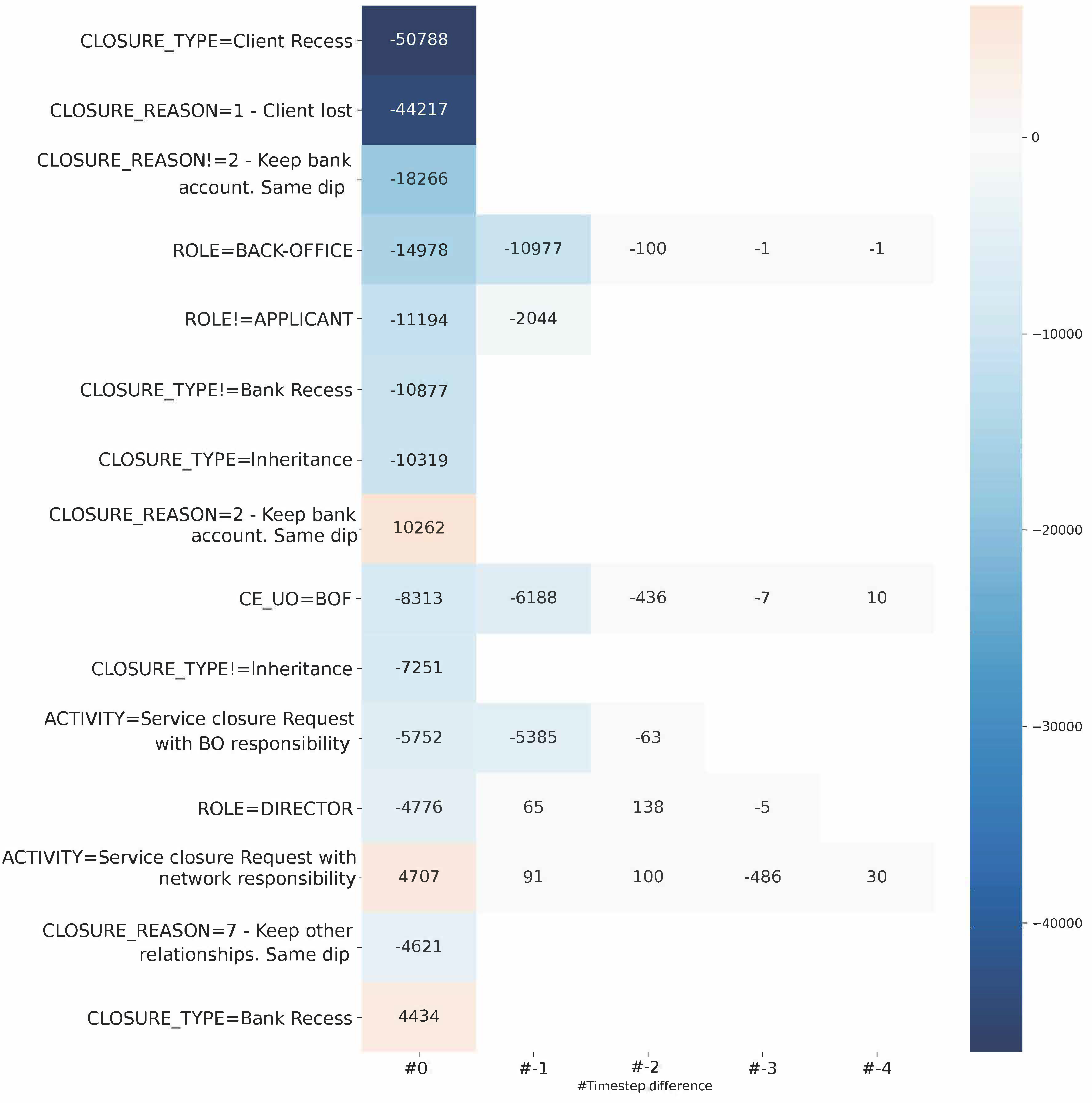}
    \caption{Offline explanations for \emph{Authorization Requested} prediction (Bank Account Closure)}
    \label{fig:Auth_Requested}
\end{figure*}

\begin{table*}[ht!]
\captionsetup{font=large}
\caption{Event logs statistics}
\footnotesize
\centering
\resizebox{\textwidth}{!}
{\begin{tabular}{|l|l|l|l|l|l|l|}
\hline
\cellcolor{LightCyan}\textbf{Event Log} &
\cellcolor{LightCyan}{\textbf{\begin{tabular}[c]{@{}c@{}}\#\\ traces\end{tabular}}} &
\cellcolor{LightCyan}{\textbf{\begin{tabular}[c]{@{}c@{}}\#\\ activities\end{tabular}}} &
\cellcolor{LightCyan}{\textbf{\begin{tabular}[c]{@{}c@{}}mean\\ events/trace\end{tabular}}} &
\cellcolor{LightCyan}{\textbf{\begin{tabular}[c]{@{}c@{}}median\\ events/trace\end{tabular}}} &
\cellcolor{LightCyan}{\textbf{\begin{tabular}[c]{@{}c@{}}mean\\ duration\end{tabular}}} &
\cellcolor{LightCyan}{\textbf{\begin{tabular}[c]{@{}c@{}}std deviation\\ duration\end{tabular}}} \\
\hline
 \textbf{Bank Account Closure} & 
 32429 & 
 15 & 
 5.5 & 
 7 & 
 15.5 days & 
 33 days  \\ 
\hline
 \cellcolor{LightGray}\textbf{BPIC 2012}  &
 \cellcolor{LightGray}12369 & 
 \cellcolor{LightGray}23 & 
 \cellcolor{LightGray}14  & 
 \cellcolor{LightGray}8 & 
 \cellcolor{LightGray}7.9 days &
 \cellcolor{LightGray}11.7 days  \\ \hline
\textbf{BPIC 2012 - W}  & 9658  & 6  & 7.5 & 6 & 11.4 days & 12.7 days  \\ \hline
\cellcolor{LightGray}\textbf{BPIC 2013}  & 
\cellcolor{LightGray}7554  & 
\cellcolor{LightGray}13 & 
\cellcolor{LightGray}8.7 & 
\cellcolor{LightGray}6 & 
\cellcolor{LightGray}12.1 days & 
\cellcolor{LightGray}28.6 days  \\  \hline
\textbf{HelpDesk 2017}  & 4580  & 14 & 4.7 & 4 & 40.9 days & 8.4 days \\ \hline
\end{tabular}}
\label{tab:dataset_statistics}
\end{table*}

\begin{table*}[ht!]
\captionsetup{font=large}
\caption{Event logs and summary statistics}
\centering
\rowcolors{3}{LightGray}{}
\begin{tabular}{ |l|c|c|c|c|c|  }

\hline
  \rowcolor{LightCyan} &
  \textbf{\begin{tabular}{@{}c@{}}Remaining time MAE \\(days) \end{tabular}}&
  \multicolumn{3}{|c|}{\textbf{Activity occurrence prediction (AUROC / APR / F1)}} &
  \textbf{Cost prediction MAE (Euro)} \\
\hline
  \textbf{Bank Account Closure} & 
  \begin{tabular}{@{}c@{}}\\\\\\ 4.37 \end{tabular}&
  \begin{tabular}{@{}c@{}}Authorization \\ Requested\\ \\ 1 / 0.99 / 0.99 \end{tabular}&
  \begin{tabular}{@{}c@{}}BO Adjustment \\ Requested\\ \\ 0.86 / 0.69 / 0.65 \end{tabular}&
  \begin{tabular}{@{}c@{}}Pending Request for \\ acquittance of heirs\\ \\ 0.99 / 0.87 / 0.90 \end{tabular}&
  \begin{tabular}{@{}c@{}}\\\\\\ 0.95 \end{tabular}\\
\hline
  \textbf{BPIC 2012} &
  {\begin{tabular}{@{}c@{}} \\\\ 6.66 \end{tabular}}&
  {\begin{tabular}{@{}c@{}}A\_ACCEPTED\\ \\ 0.92 / 0.66 / 0.60\end{tabular}} &
  {\begin{tabular}{@{}c@{}}A\_CANCELLED\\ \\ 0.74 / 0.52 / 0.37\end{tabular}} &
  {\begin{tabular}{@{}c@{}}A\_DECLINED\\ \\ 0.77 / 0.55 / 0.51\end{tabular}} &
  - \\
\hline
\textbf{BPIC 2012 - W} &
 \begin{tabular}{@{}c@{}} \\\\ 7.84 \end{tabular}&
  - &
  - &
  - &
  - \\
\hline
  \textbf{BPIC 2013} &
  {\begin{tabular}{@{}c@{}} \\\\ 11.82 \end{tabular}}&
  {\begin{tabular}{@{}c@{}}Wait - User\\ \\ 0.50 / 0.43 / 0.45\end{tabular}} &
  {\begin{tabular}{@{}c@{}}2nd / 3rd line\\ \\ 0.90 / 0.90 / 0.81\end{tabular}} &
  - &
  - \\
\hline  
\textbf{HelpDesk 2017} &
 \begin{tabular}{@{}c@{}} \\\\ 5.96 \end{tabular}&
  - &
  - &
  - &
  - \\
\hline
\end{tabular}
\label{tab:results}
\end{table*}

We show here the application of our framework to other two activities that are of interest for the bank, \textit{Autorization Required} and \textit{Pending Request for Acquittance of Heirs}.

In particular, in Figure \ref{fig:Auth_Requested} are shown the explanations related to the  \emph{Authorization Requested} prediction. Here, most of the times the LSTM is predicting that the activity will not be performed (as it can be seen by the negative values), and this is correct since the authorization to proceed further is usually requested to a director in the early stages of the process, rather than at the end. 
In particular, activities performed by the resource BOF (value -8313 in the heatmap) or by resources with role Back-office (value -14978) are all performed after \emph{Authorization Requested}; this is of course also true when these activities have been performed in previous steps (see in particular columns related to timestep difference -1 and -2).
Conversely, when a customer does not want to change the bank but wants just to close one of its bank accounts among different ones he owns (labeled as \emph{Closure\_Reason=2 - Keep bank account. Same dip}), most of the times a director's authorization need to be requested (value 10262); on the contrary, when a customer decides to change the bank and close definitively its bank account (labeled by \emph{Closure\_type=Client Recess} and by \emph{Closure\_Reason=1 - Client lost}), then a director's authorization is not needed.

\begin{figure*}[t!]
    \centering
    \includegraphics[width=1 \textwidth]{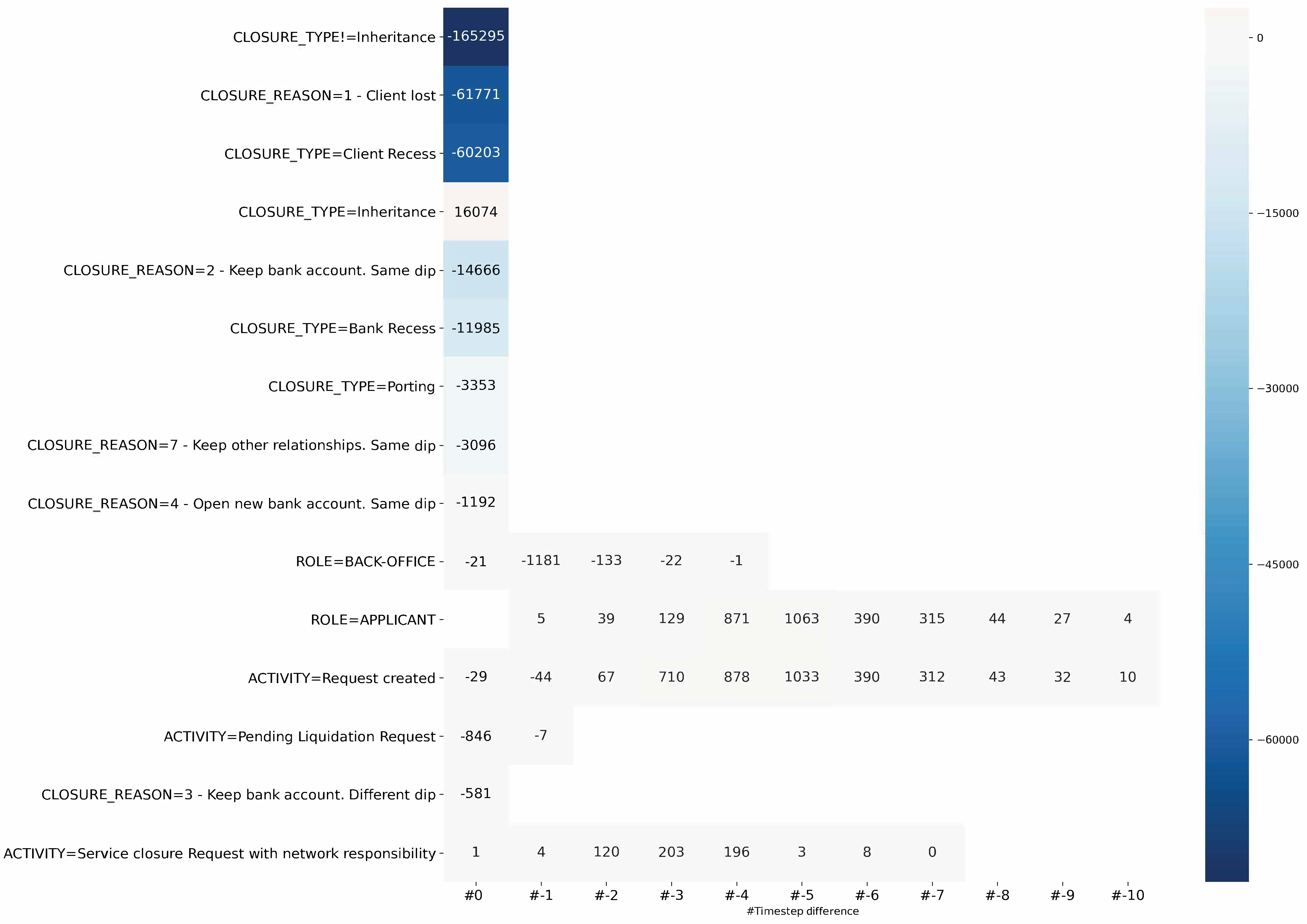}
    \caption{Offline explanations for \emph{Pending Request for acquittance of heirs} prediction (Bank Account Closure)}
    \label{fig:Pending Request for acquittance of heirs}
\end{figure*}

\begin{figure*}[t!]
    \centering
    \includegraphics[width=1 \textwidth]{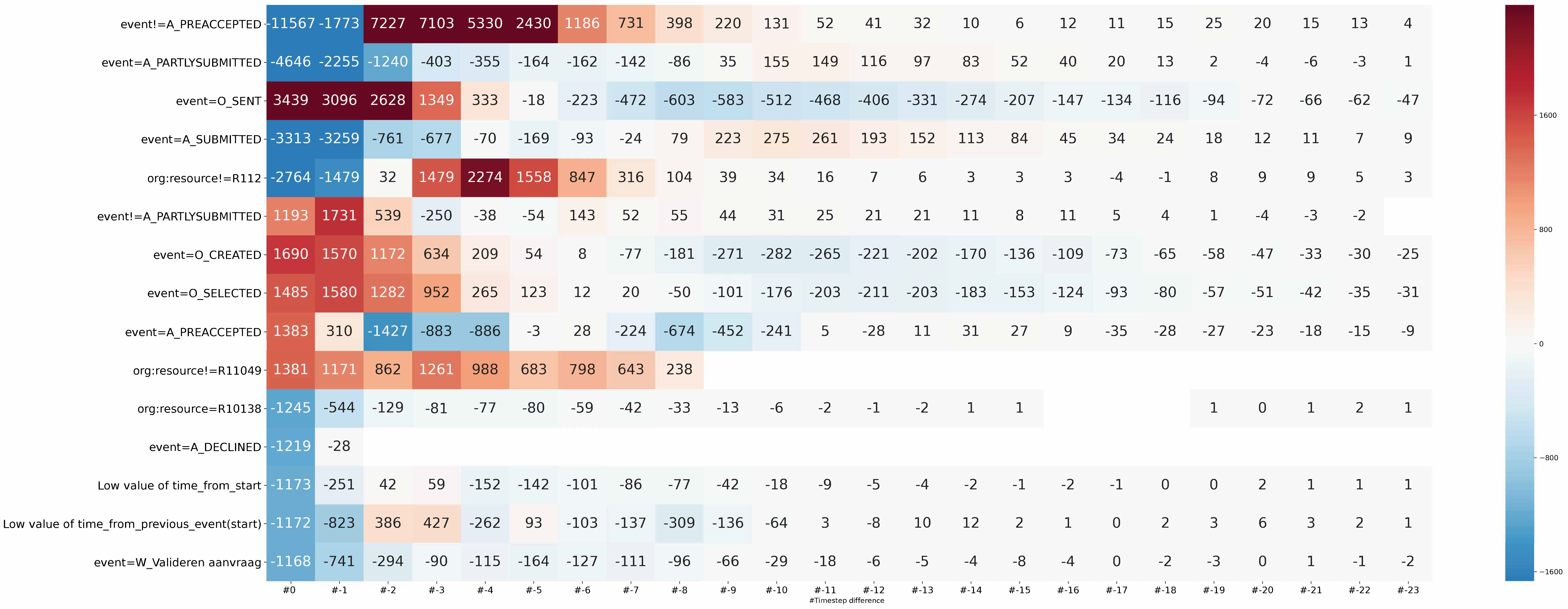}
    \caption{Offline explanations for BPI12 remaining time prediction}
    \label{fig:bpi12_time}
\end{figure*}

\begin{figure*}[t!]
    \centering
    \includegraphics[width=1 \textwidth]{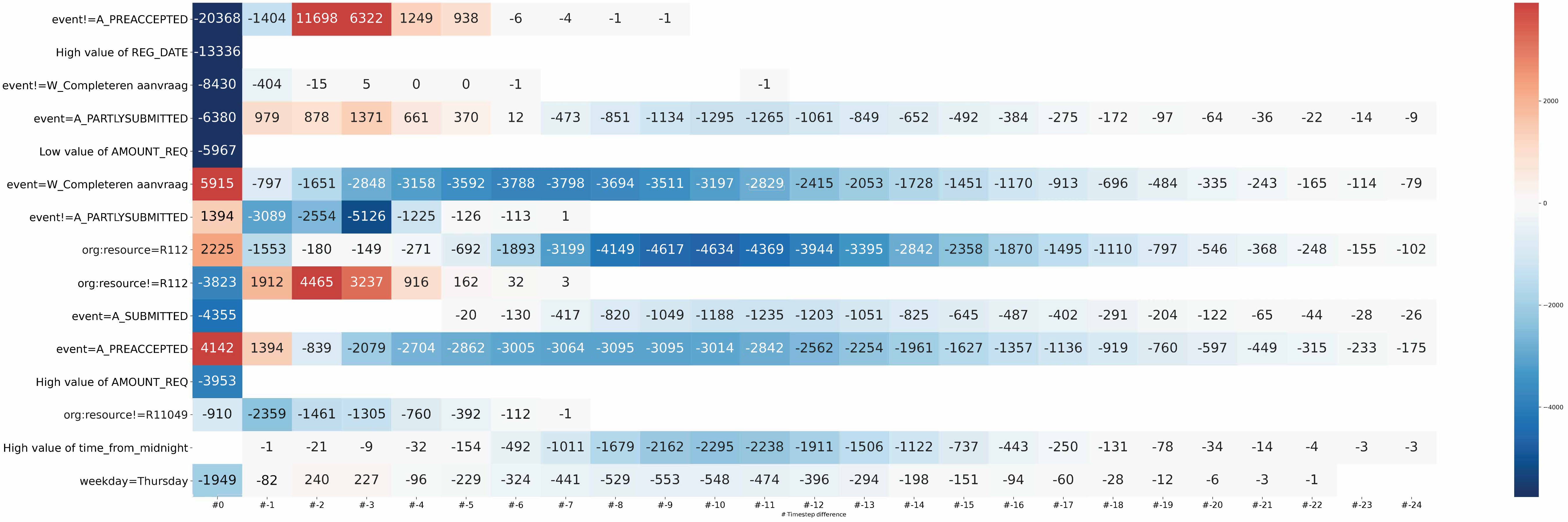}
    \caption{Offline explanations for BPI12 \emph{A\_ACCEPTED} prediction}
    \label{fig:bpi12_accepted}
\end{figure*}

\begin{figure*}[t!]
    \centering
    \includegraphics[width=1 \textwidth]{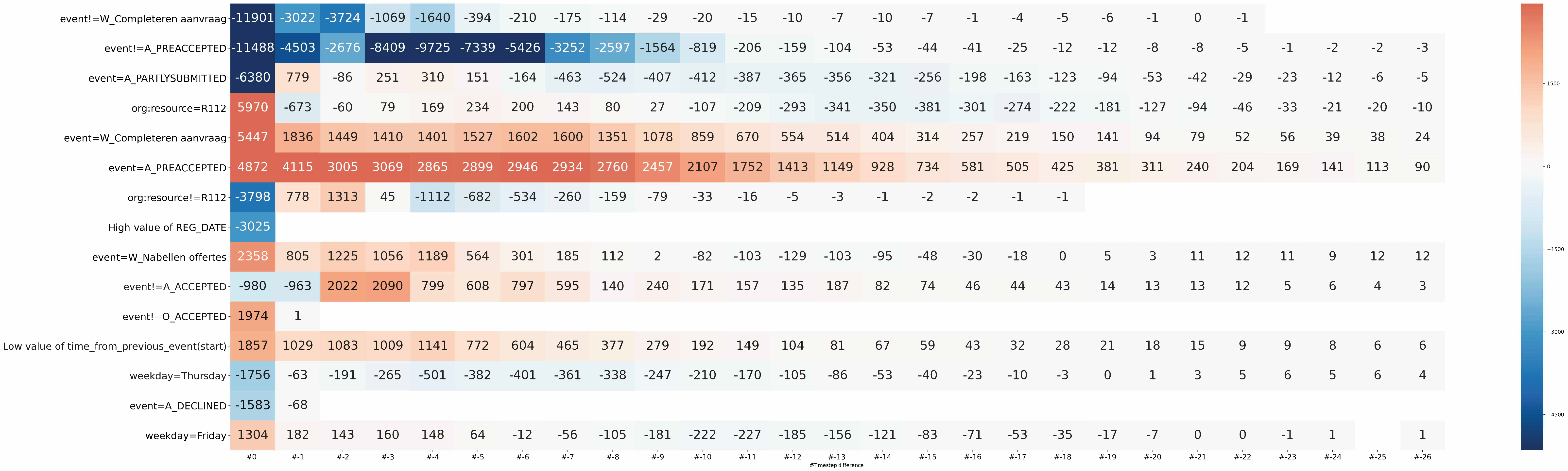}
    \caption{Offline explanations for BPI12 \emph{A\_CANCELLED} prediction}
    \label{fig:bpi12_cancelled}
\end{figure*}

\begin{figure*}[t!]
    \centering
    \includegraphics[width=1 \textwidth]{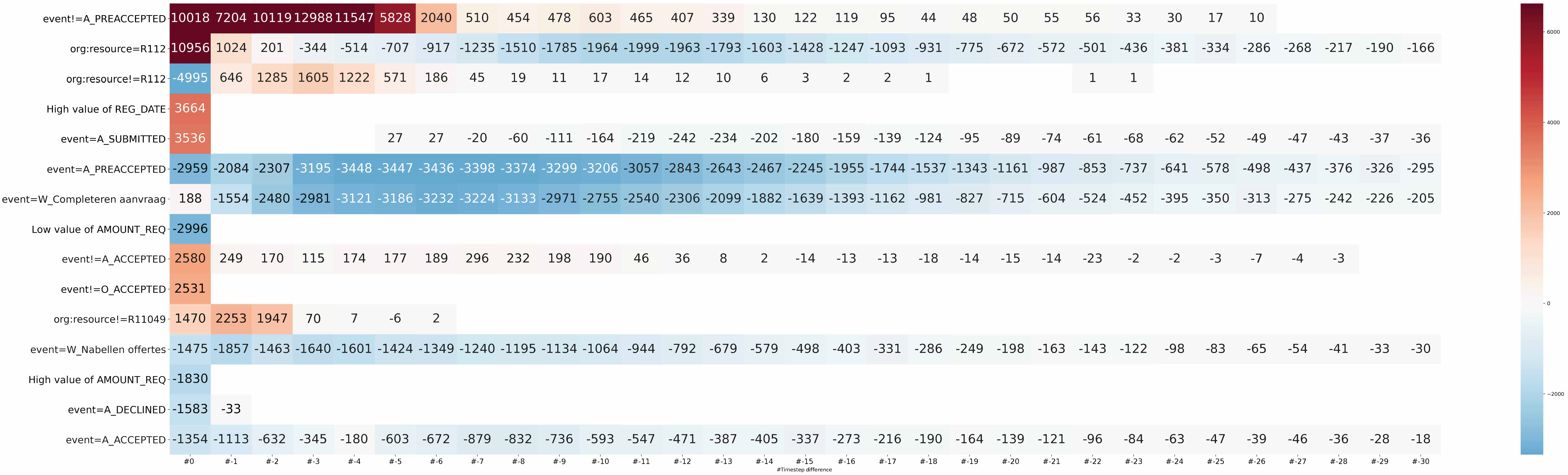}
    \caption{Offline explanations for BPI12 \emph{A\_DECLINED} prediction}
    \label{fig:bpi12_declined}
\end{figure*}

\subsection{BPIC 2012}

This is the dataset from BPI 2012 challenge, and represents a real-life log of Dutch Financial Institute.
Figure \ref{fig:bpi12_time} shows the application for the remaining time prediction in the bpi12 dataset.
The largest value is represented by \emph{event!=A\_PREACCEPTED}, meaning that this activity is particularly important for the model. As a matter of fact, when this activity is currently performed (\emph{event=A\_PREACCEPTED} with value 1383 and timestep difference 0), the offer is more likely to be accepted by the customer, leading to the acceptance process, which takes longer to conclude compared to when the offer is declined (when \emph{event=A\_DECLINED} the value is -1219, so a smaller remaining time is expected). 
The explanation \emph{event!=A\_PREACCEPTED} with timestep differences 0 and -1 (values respectively -11567 and -1773) is mainly found in the early stages of the process; as a matter of fact, not until is this activity performed, there are high probabilities that the offer could still be declined, leading to a much shorter process, with the model correctly predicting a smaller remaining time. Conversely, lower timestep differences (-2, -3, -4, -5) indicate that the process is currently in the acceptance path and, since it takes longer to complete, a larger remaining time is predicted. This is also indicated by the activities \emph{O\_SENT}, \emph{O\_CREATED}, and \emph{O\_SELECTED}; when they are currently performed (timestep difference 0), or when they have been previously performed (see columns related to timestep differences -1, -2, -3 and -4), the prediction of the remaining time is increased. Note that the influence of these 3 activities decreases with the passing of time, becoming negative at a certain point: the case is getting closer to the end. 
Finally, states of the work item belonging to the application seems to be not relevant for the prediction.
The financial institute aims also to identify in advance if a loan will be declined (represented by the activity \emph{A\_DECLINED}). The heatmap related to \emph{A\_DECLINED} prediction (Figure \ref{fig:bpi12_declined}) shows again that the activity \emph{A\_PREACCEPTED} being performed or not is the most influential factor. When \emph{A\_PREACCEPTED} is not currently performed or it has not been performed in the past (as it can be seen in the positive values in the first row of the heatmap), it is likely that the offer is going to be declined; on the contrary, when \emph{A\_PREACCEPTED} is performed or when it has been performed in the past (as it can be seen by the negative values in the row labeled as \emph{event=A\_PREACCEPTED}), then it is very probable that the offer will be accepted; consequently, it is predicted that the offer will not be declined.
In general when information for the application have been filled (labeled as \emph{event=W\_Completeren aanvraag}), \emph{A\_DECLINED} is predicted to not happen (shown in the columns related to timestep difference -1 or lower), but when information have just been filled (as it can be seen by the value 188 in the timestep difference column 0, which refers to the last event of the prefix) there is still a little probability that the offer is going to be declined. Finally, when the amount of the requested loan is low (labeled as \emph{Low value of AMOUNT\_REQ} it is less probable that the loan is going to be declined by the institute (indicated by the negative value -2996) compared to when the requested amount is high (labeled as \emph{High value of AMOUNT\_REQ}).

\subsection{BPIC 2012-W}

\begin{figure*}[t!]
    \centering
    \includegraphics[width=1 \textwidth]{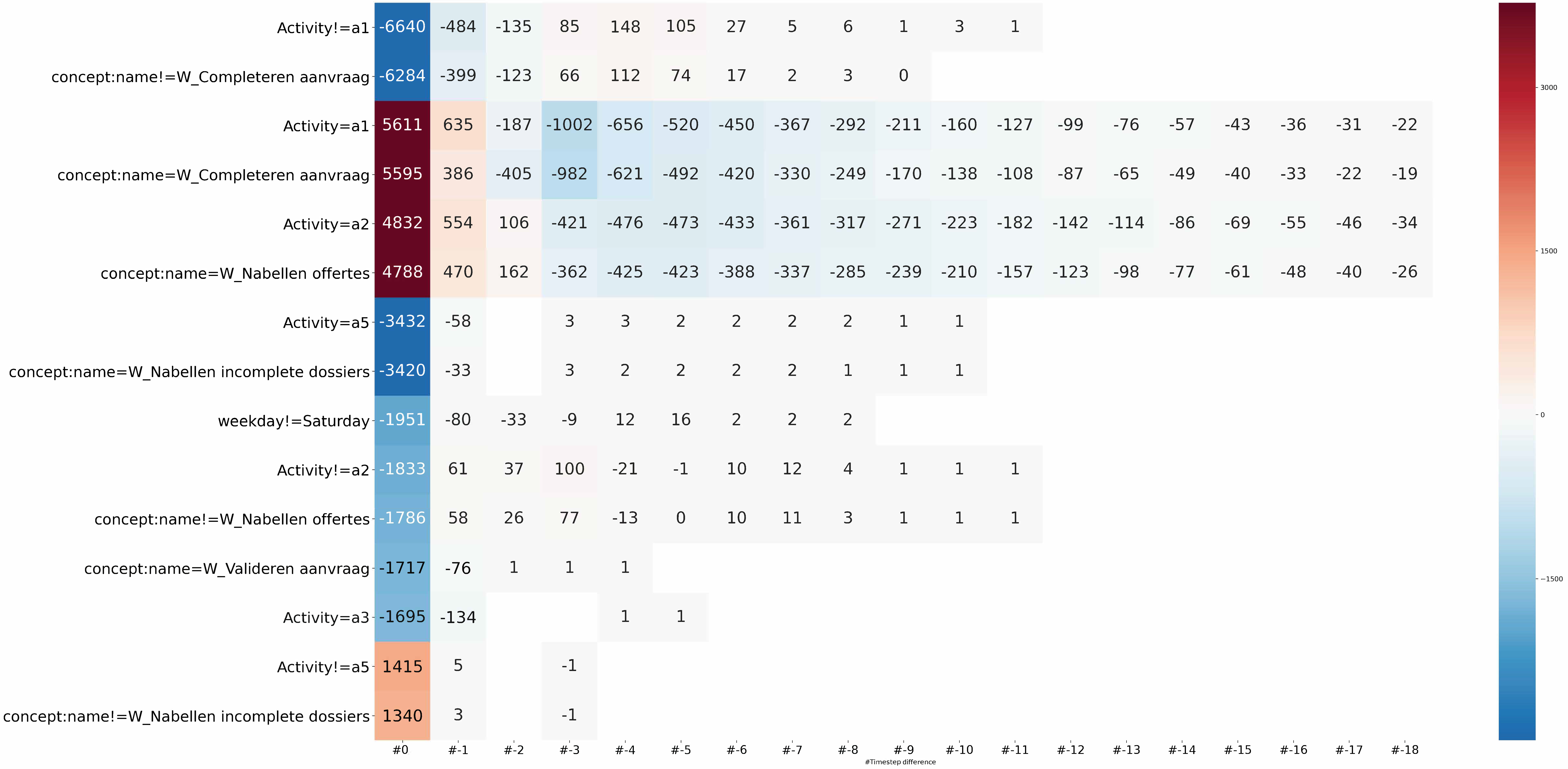}
    \caption{Offline explanations for BPI12-W remaining time prediction}
    \label{fig:bpi12_W_time}
\end{figure*}

Figure \ref{fig:bpi12_W_time} shows the application for the remaining time prediction in the bpi12-W dataset, that is the dataset derived from bpi12 challenge, containing only the states of the work items belonging to the application. Here the largest and most important values are represented by the activity attribute; please note that here \emph{Activity} and \emph{concept:name} refer to the same thing (and this is also correctly understood by the model, which gives them the same importance), so for simplicity we will address them as activity. 
In particular, \emph{W\_Completeren aanvraag} and \emph{W\_Nabellen offertes} (labeled respectively as \emph{concept:name=W\_Completeren aanvraag} and \emph{concept:name=W\_Nabellen offertes}) are always performed in the initial part of cases, and, hence, when they are currently performed (timestep difference 0) an higher remaining time is predicted. Instead, when these two activities have been performed in the past it means that either they are still performed because of internal inefficiencies that cause reworks or other activities are currently performed; in any case the process is not anymore in its early stages and hence a smaller remaining time is predicted (this can be noticed in the heatmap looking at the columns referring to lower timestep differences for those two attributes).   
As a matter of fact, when these two activities are not currently performed (labeled as \emph{concept:name!=W\_Completeren aanvraag} and \emph{concept:name!=W\_Nabellen offertes}) it's even more influencial in predicting a smaller remaining time, because those inefficient reworks have already been done.
According to the same logic, \emph{W\_Valideren aanvraag} and \emph{W\_Nabellen incomplete dossiers} are performed in the final part of cases and when they occur (labeled as \emph{concept:name=W\_Valideren aanvraag} and \emph{concept:name=W\_Nabellen incomplete dossiers}) a smaller remaining time is correctly predicted.

\subsection{BPIC 2013}

\begin{figure*}[t!]
    \centering
    \includegraphics[width=1 \textwidth]{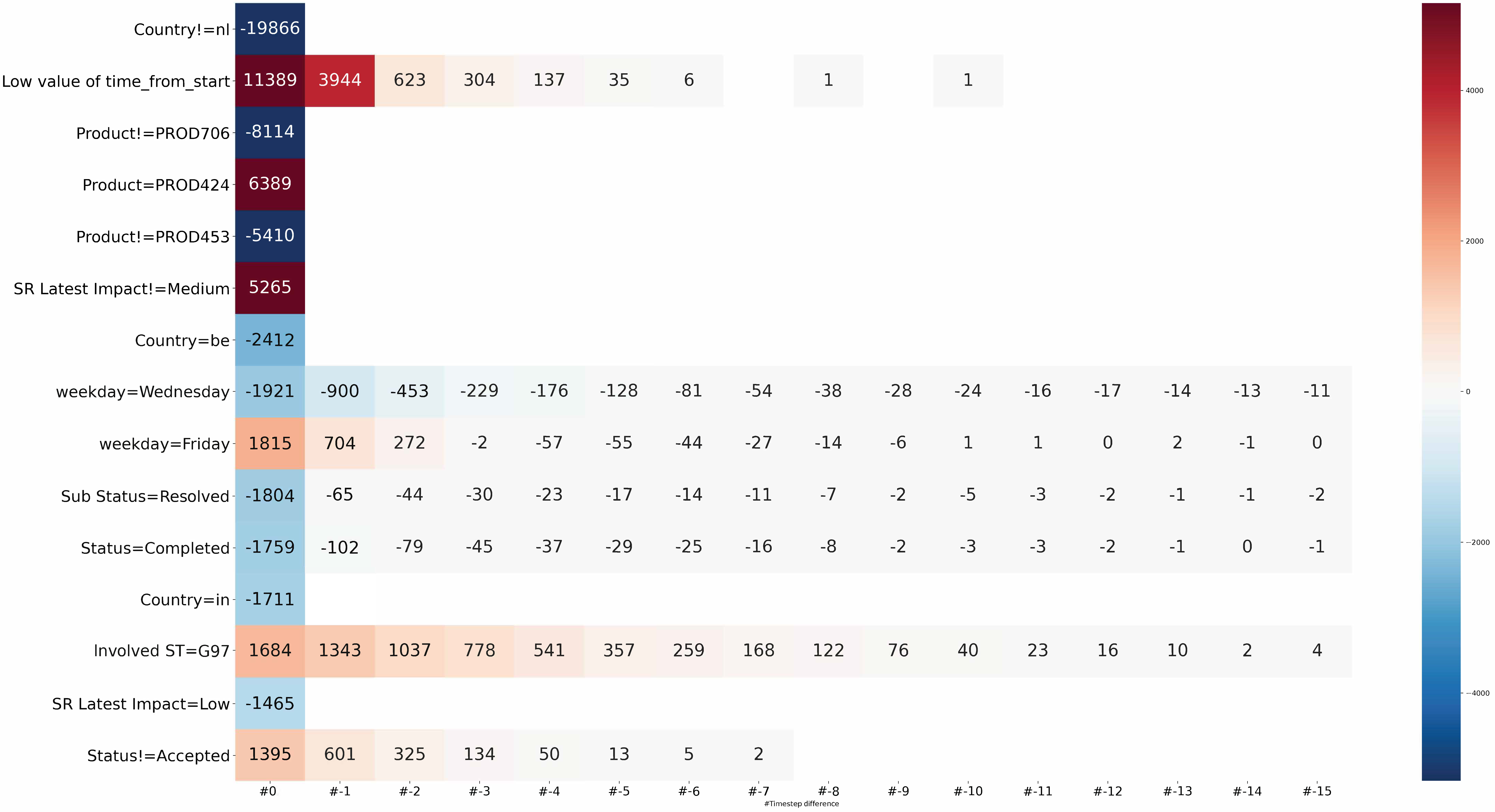}
    \caption{Offline explanations for BPI13 remaining time prediction}
    \label{fig:bpi13_time}
\end{figure*}

\begin{figure*}[t!]
    \centering
    \includegraphics[width=1 \textwidth]{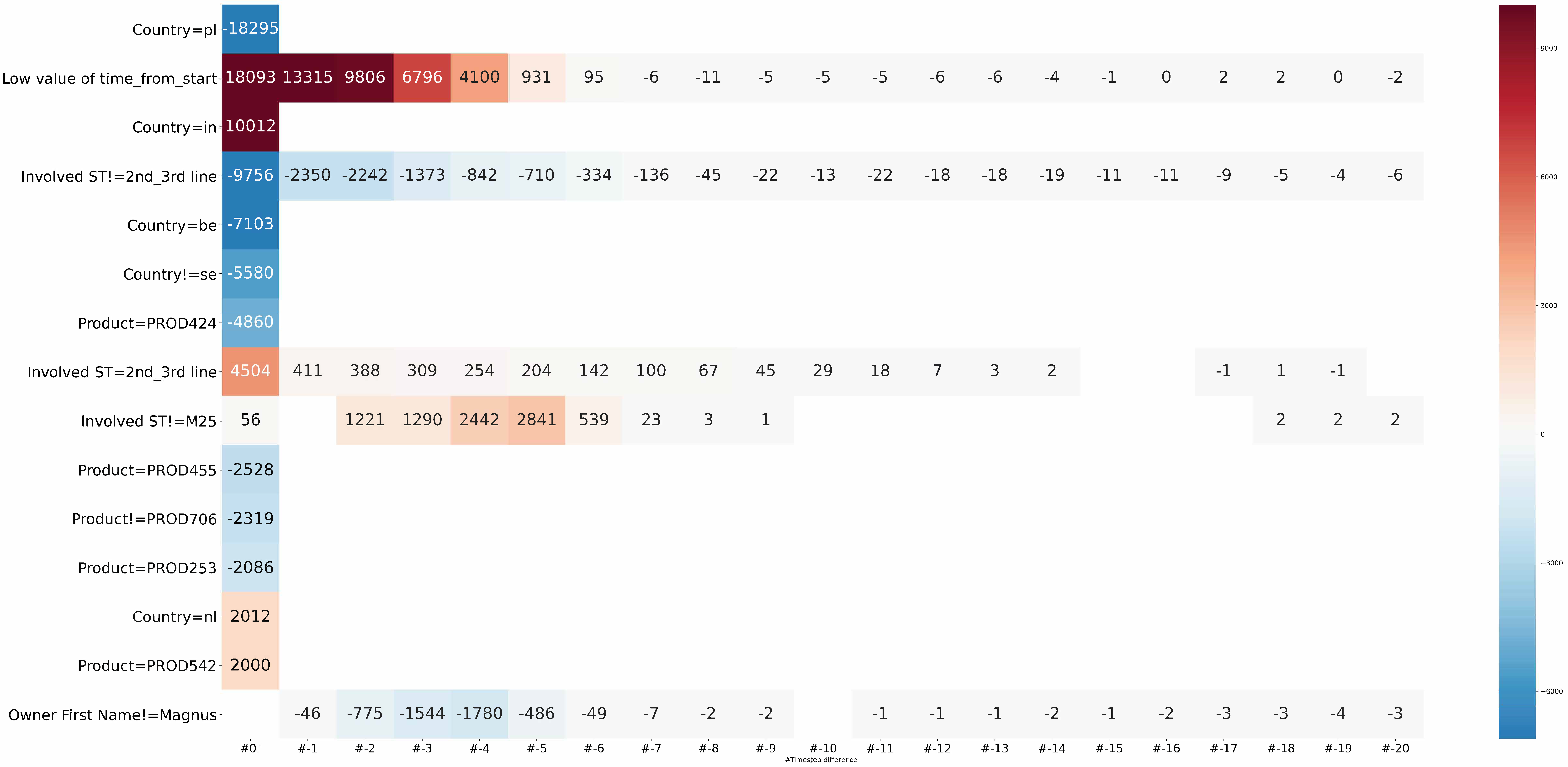}
    \caption{Offline explanations for BPI13 push to front line prediction}
    \label{fig:bpi13_push_to_front}
\end{figure*}

\begin{figure*}[t!]
    \centering
    \includegraphics[width=1 \textwidth]{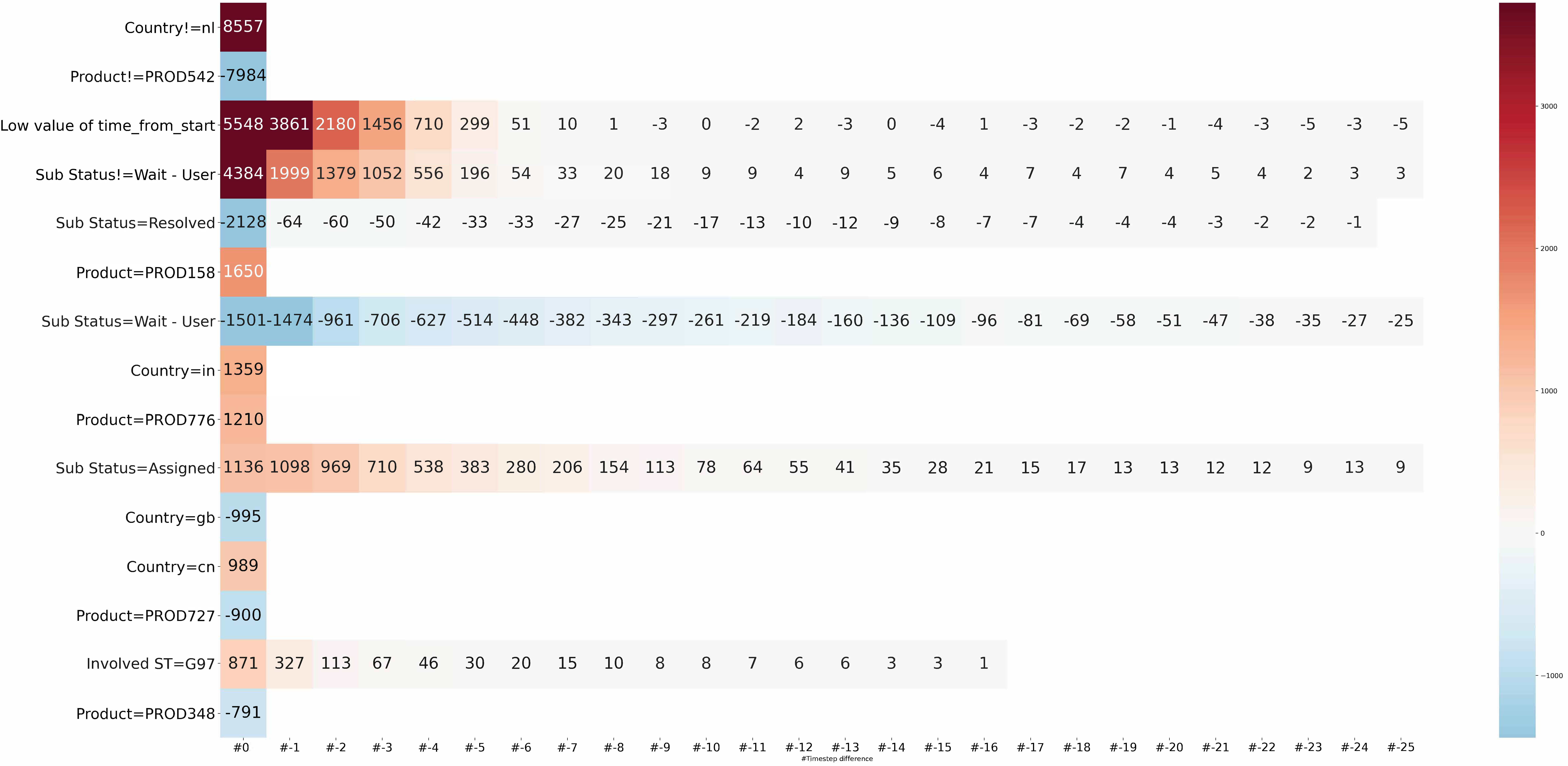}
    \caption{Offline explanations for BPI13 wait\_user prediction}
    \label{fig:bpi13_wait_user}
\end{figure*}

\begin{figure*}[ht!]
    \centering
    \includegraphics[width=1 \textwidth]{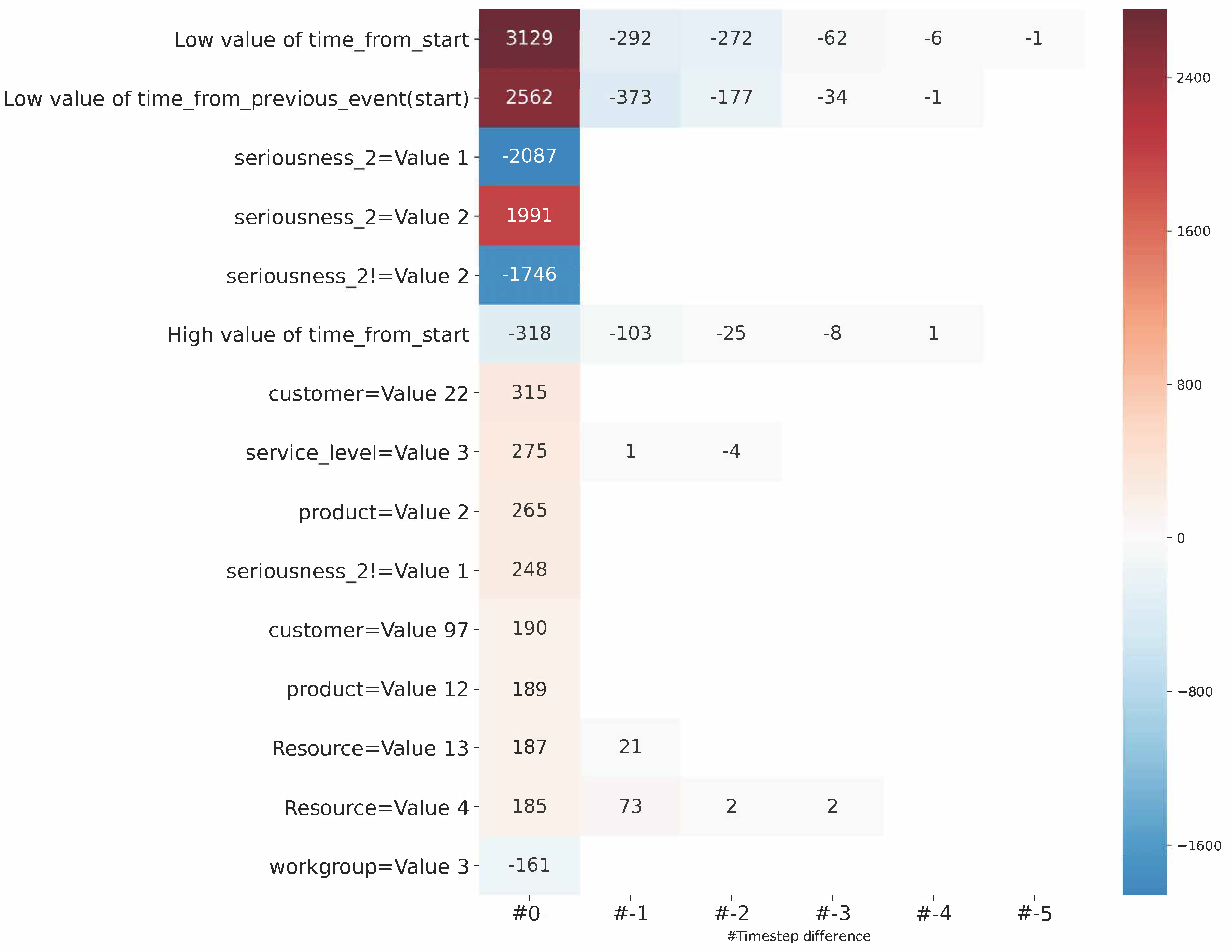}
    \caption{Offline explanations for HelpDesk17 remaining time prediction}
    \label{fig:helpdesk17_time}
\end{figure*}

This is a dataset from BPI 2013 challenge, extracted from a company's incident management system.
One question deals with the strategy of the company that most of the incidents need to be resolved by the first line support teams without involving 2nd or 3rd support line teams (push to front), leading to a more efficient process, so here we focused on predicting if at least one activity is going to be performed by a resource not belonging to the first line support team. The heatmap related to push to front prediction is shown in Figure \ref{fig:bpi13_push_to_front}; the fact that the country in which the problem has been addressed was Poland (labeled as \emph{Country=pl}) is the most influential. The negative value -18295 means that the involvement of a 2nd or 3rd line resource is unlikely to happen. A further analysis of the data confirms this finding: when the country is Poland, then a 2nd / 3rd line resource is involved in the 18\% of cases.
Values -4860 when the product is 424 (\emph{Product=PROD424}) and -2528 when the product is 455 (\emph{Product=PROD455}) indicate as well that the involvement of a resource not belonging to the first line support team is unlikely to happen.
Conversely, when the country is India (labeled as \emph{Country=in} with the value 10012), then an activity performed by a 2nd / 3rd line resource is more likely to happen; again an analysis of the data confirms that when the country is India a 2nd / 3rd line resource is involved only in the 79\% of cases.
Other important attributes positively contributing to predict an involvement of a resource not from the first line support team are related to the fact that the process is in the initial stages (labeled as \emph{Low value of time from start} with the value 18093 in timestep difference 0 column) and that an activity is currently performed by a resource belonging to the 2nd or 3rd line support team (labeled as \emph{Involved ST=2nd\_3rd line} with the value 4504); note that the influence of these attributes decreases with the passing of time (see columns related to timestep difference -1, -2, -3 and so on).

The company has also an interest in understanding if people working in the company is abusing the \emph{Wait - User} substatus to hide inefficiencies in the process that would otherwise being detected by KPIs measuring the total resolution time of an incident. Figure \ref{fig:bpi13_wait_user} shows the results for the \emph{Wait - User} substatus prediction.
Again, if the process is in the initial stages, the \emph{Wait - User} substatus is of course more likely to happen, and this importance decreases with time; on the other side, if a \emph{Wait - User} substatus is currently being performed (labeled as \emph{Sub Status=Wait - User}) or has been performed in the past (see columns related to negative timestep differences), then this activity is not predicted to be performed again.
This substatus is similarly unlikely to happen if \emph{Sub Status=Resolved} is currently performed, since this is the activity that is set when the incident has already been solved.
Other important attributes are the country in which the problem has been addressed and the product that presented a problem or which had an incident; as an example, if the product is 158 (labeled as \emph{Product=PROD158} with value 1650), 776 (labeled as \emph{Product=PROD776} with value 1210) or the country is India (labeled as \emph{Country=in} with value 1359), then a \emph{Wait - User} activity is likely to happen. A further analysis of the data confirms this finding: when the product is 158, 778, or the country is India, a \emph{Wait - User} has been used respectively in the 85\%, 66\%, and 63\% of cases.  

\subsection{HelpDesk 2017}

This dataset is a real-life log of SIAV s.p.a. company in Italy, and represents instances of a ticketing process in the company helpdesk area.
Figure \ref{fig:helpdesk17_time} refers to the application for the remaining time prediction.
Here, again, the fact that the process is in the early stages is the most influential factor. The positive value at timestep difference 0 (3129) indicates that the influence is towards increasing the value, namely towards having a largest remaining time. Other two important factors are related to the seriousness and priority of the ticket; if the seriousness is marked as 1 (labeled as \emph{seriousness\_2=Value 1} with value -2087), then a smaller remaining time is predicted, while if the seriousness is marked as 2 (labeled as \emph{seriousness\_2=Value 2} with value 1991), a larger remaining time is indicated.
This fact is shown as well in the data, since when the seriousness of the ticket is marked as 1, the total time for the case to be executed is in general 6 days more compared to when the seriousness is marked as 2.

\noindent \textbf{Acknowledgement.} This work is supported by MINECO and FEDER funds under grant TIN2017-86727-C2-1-R and by the Department of Mathematics, University of Padua, with HPC resources.

\bibliographystyle{splncs04}
\bibliography{main.bib}

\begin{thebibliography}{10}
\providecommand{\url}[1]{\texttt{#1}}
\providecommand{\urlprefix}{URL }
\providecommand{\doi}[1]{https://doi.org/#1}

\bibitem{alvarez2018robustness}
Alvarez-Melis, D., Jaakkola, T.S.: On the robustness of interpretability
  methods. arXiv preprint arXiv:1806.08049  (2018)

\bibitem{DBLP:conf/iclr/2015}
Bahdanau, D., Cho, K., Bengio, Y.: Neural machine translation by jointly
  learning to align and translate. In: The 3rd International Conference on
  Learning Representations, {ICLR} 2015, San Diego, CA, USA, May 7-9, 2015,
  Conference Track Proceedings (2015)

\bibitem{10.1007/978-3-319-15895-2_46}
Breuker, D., Delfmann, P., Matzner, M., Becker, J.: Designing and evaluating an
  interpretable predictive modeling technique for business processes. In:
  Business Process Management Workshops. pp. 541--553. Springer (2015)

\bibitem{doshivelez2017rigorous}
Doshi-Velez, F., Kim, B.: Towards a rigorous science of interpretable machine
  learning (2017)

\bibitem{Hochreiter1997}
Hochreiter, S., {Urgen Schmidhuber}, J.: {Long Short-Term Memory}. Neural
  Computation  \textbf{9}(8),  1735--1780 (1997)

\bibitem{shap}
Lundberg, S.M., Lee, S.I.: A unified approach to interpreting model
  predictions. In: Advances in neural information processing systems. pp.
  4765--4774 (2017)

\bibitem{lundberg2018explainable}
Lundberg, S.M., Nair, B., Vavilala, M.S., Horibe, M., Eisses, M.J., Adams, T.,
  Liston, D.E., Low, D.K.W., Newman, S.F., Kim, J., et~al.: Explainable
  machine-learning predictions for the prevention of hypoxaemia during surgery.
  Nature biomedical engineering  \textbf{2}(10), ~749 (2018)

\bibitem{Marquez-Chamorro18}
M{\'{a}}rquez{-}Chamorro, A.E., Resinas, M., Ruiz{-}Cort{\'{e}}s, A.:
  Predictive monitoring of business processes: {A} survey. {IEEE} Transaction
  on Services Computing  \textbf{11}(6),  962--977 (2018)

\bibitem{inbook}
Meacham, S., Isaac, G., Nauck, D., Virginas, B.: Towards Explainable AI: Design
  and Development for Explanation of Machine Learning Predictions for a Patient
  Readmittance Medical Application, pp. 939--955 (06 2019)

\bibitem{molnar2019}
Molnar, C.: Interpretable Machine Learning (2020)

\bibitem{LSTM_time}
Navarin, N., Vincenzi, B., Polato, M., Sperduti, A.: {LSTM} networks for
  data-aware remaining time prediction of business process instances. In:
  Proceedings of the {IEEE} Symposium Series on Computational Intelligence
  {(SSCI 2017)} (2017)

\bibitem{10.1007/s11257-017-9195-0}
Nunes, I., Jannach, D.: A systematic review and taxonomy of explanations in
  decision support and recommender systems. User Modeling and User-Adapted
  Interaction  \textbf{27}(3–5),  393–444 (Dec 2017)

\bibitem{Park19}
{Park}, G., {Song}, M.: Prediction-based resource allocation using lstm and
  minimum cost and maximum flow algorithm. In: International Conference on
  Process Mining (ICPM). pp. 121--128 (2019)

\bibitem{Rehse2019}
Rehse, J.R., Mehdiyev, N., Fettke, P.: Towards explainable process predictions
  for industry 4.0 in the dfki-smart-lego-factory. KI - K{\"u}nstliche
  Intelligenz  \textbf{33}(2),  181--187 (Jun 2019)

\bibitem{lime}
Ribeiro, M.T., Singh, S., Guestrin, C.: ``why should {I} trust you'':
  Explaining the predictions of any classifier. In: Proceedings of the 22nd
  {ACM} {SIGKDD} International Conference on Knowledge Discovery and Data
  Mining, San Francisco. pp. 1135--1144 (2016)

\bibitem{DBLP:conf/naacl/JainW19}
Sarthak, J., Wallace, B.C.: Attention is not explanation. In: Proceedings of
  the 2019 Conference of the North American Chapter of the Association for
  Computational Linguistics: Human Language Technologies. pp. 3543--3556.
  Association for Computational Linguistics (2019)

\bibitem{serrano-smith-2019-attention}
Serrano, S., Smith, N.A.: Is attention interpretable? In: Proceedings of the
  57th Annual Meeting of the Association for Computational Linguistics. pp.
  2931--2951. Association for Computational Linguistics, Florence, Italy (2019)

\bibitem{shapley1953value}
Shapley, L.S.: A value for n-person games. Contributions to the Theory of Games
   \textbf{2}(28),  307--317 (1953)

\bibitem{shrikumar2017learning}
Shrikumar, A., Greenside, P., Kundaje, A.: Learning important features through
  propagating activation differences. In: Proceedings of the 34th International
  Conference on Machine Learning-Volume 70. pp. 3145--3153. JMLR. org (2017)

\bibitem{Shu:2019:DEF:3292500.3330935}
Shu, K., Cui, L., Wang, S., Lee, D., Liu, H.: Defend: Explainable fake news
  detection. In: International Conference on Knowledge Discovery \& Data
  Mining, SIGKDD. pp. 395--405. ACM (2019)

\bibitem{gradients}
Sundararajan, M., Taly, A., Yan, Q.: Axiomatic attribution for deep networks.
  In: Proceedings of the 34th International Conference on Machine
  Learning-Volume 70. pp. 3319--3328. JMLR. org (2017)

\bibitem{TaxVRD17}
Tax, N., Verenich, I., La~Rosa, M., Dumas, M.: Predictive business process
  monitoring with {LSTM} neural networks. In: Proceedings of 29th International
  Conference on Advanced Information Systems Engineering (CAiSE 2017). pp.
  477--492 (2017)

\end{thebibliography}

\end{document}